\documentclass[sigconf]{acmart} % kdd recommended
\usepackage{multirow}
\usepackage{xcolor}
\usepackage{colortbl}
\usepackage{subcaption} 
\usepackage{xcolor}
\usepackage{fancyvrb}
\usepackage{verbatim}
\usepackage{framed}
\usepackage{listings}
\usepackage{enumitem}
\AtBeginDocument{%
  }

\setcopyright{acmlicensed}
\copyrightyear{2026}
\acmYear{2026}
\acmDOI{XXXXXXX.XXXXXXX}
\acmConference[KDD '26]{The 32nd ACM SIGKDD Conference on Knowledge Discovery and Data Mining V.2}{August 09--13, 2026}{Jeju Island, Republic of Korea}
\acmISBN{978-1-4503-XXXX-X/2018/06}

\newenvironment{promptbox}[2][gray!75!black]{%
  \def\FrameCommand{\fcolorbox{#1}{white}}%
  \setlength{\FrameSep}{0pt}%
  \MakeFramed{\advance\hsize-\width \FrameRestore}%
  \noindent\colorbox{#1}{\parbox{\dimexpr\linewidth-2\fboxsep\relax}{\textcolor{white}{\textbf{\small #2}}}}\par%
  
  \nointerlineskip%
  \small\ttfamily%
}{%
  \endMakeFramed%
}

\begin{document}

\title{SAGE: A Service Agent Graph-guided Evaluation Benchmark}

% \author{Ling Shi}
% \affiliation{%

\author{Ling Shi}
\authornote{Both authors contributed equally to this research.}
\affiliation{%
  \institution{Tianjin University}
  \city{Tianjin}
  \country{China}}

\author{Yuqin Dai}
\authornotemark[1]
\affiliation{%
  \institution{Tsinghua University}
  \city{Beijing}
  \country{China}}

\author{Ziyin Wang}
\affiliation{%
  \institution{Tianjin University}
  \city{Tianjin}
  \country{China}}

\author{Ning Gao}
\affiliation{%
  \institution{Beihang University}
  \city{Beijing}
  \country{China}}

\author{Wei Zhang}
\affiliation{%
  \institution{Beijing University of Posts and Telecommunications}
  \city{Beijing}
  \country{China}}

\author{Chaozheng Wang}
\affiliation{%
  \institution{The Chinese University of Hong Kong}
  \city{Hong Kong}
  \country{China}}

\author{Yujie Wang}
\affiliation{%
  \institution{Independent Researcher}
  \city{Beijing}
  \country{China}}

\author{Wei He}
\affiliation{%
  \institution{Independent Researcher}
  \city{Beijing}
  \country{China}}

\author{Jinpeng Wang}
\affiliation{%
  \institution{Independent Researcher}
  \city{Beijing}
  \country{China}}

\author{Deyi Xiong}
\authornote{Corresponding Author}
\affiliation{%
  \institution{Tianjin University}
  \city{Tianjin}
  \country{China}}

\renewcommand{\shortauthors}{Ling Shi et al.}


\begin{abstract}

The development of Large Language Models (LLMs) has catalyzed automation in customer service, yet benchmarking their performance remains challenging. Existing benchmarks predominantly rely on static paradigms and single-dimensional metrics, failing to account for diverse user behaviors or the strict adherence to structured Standard Operating Procedures (SOPs) required in real-world deployments. To bridge this gap, we propose \textbf{SAGE} (\textbf{S}ervice \textbf{A}gent \textbf{G}raph-guided \textbf{E}valuation), a universal multi-agent benchmark for automated, dual-axis assessment. SAGE formalizes unstructured SOPs into Dynamic Dialogue Graphs, enabling precise verification of logical compliance and comprehensive path coverage. We introduce an Adversarial Intent Taxonomy and a modular Extension Mechanism, enabling low-cost deployment across domains and facilitating automated dialogue data synthesis. Evaluation is conducted via a framework where Judge Agents and a Rule Engine analyze interactions between User and Service Agents to generate deterministic ground truth. Extensive experiments on 27 LLMs across 6 industrial scenarios reveal a significant ``Execution Gap'' where models accurately classify intents but fail to derive correct subsequent actions. We also observe ``Empathy Resilience'', a phenomenon where models maintain polite conversational facades despite underlying logical failures under high adversarial intensity.
\textbf{Code and resources are available at \url{https://anonymous.4open.science/r/SAGE-Bench-4CD3/}}.
\end{abstract}
\begin{CCSXML}
<ccs2012>
   <concept>
       <concept_id>10010147.10010178.10010199.10010202</concept_id>
       <concept_desc>Computing methodologies~Multi-agent planning</concept_desc>
       <concept_significance>500</concept_significance>
       </concept>
   <concept>
       <concept_id>10010147.10010178.10010179.10010181</concept_id>
       <concept_desc>Computing methodologies~Discourse, dialogue and pragmatics</concept_desc>
       <concept_significance>500</concept_significance>
       </concept>
 </ccs2012>
\end{CCSXML}

\ccsdesc[500]{Computing methodologies~Multi-agent planning}
\ccsdesc[500]{Computing methodologies~Discourse, dialogue and pragmatics}


\keywords{Service Agent, Graph-guided Evaluation, Multi-agent Interaction}


\maketitle

\section{Introduction}
\begin{figure}
    \centering
    \includegraphics[width=\linewidth]{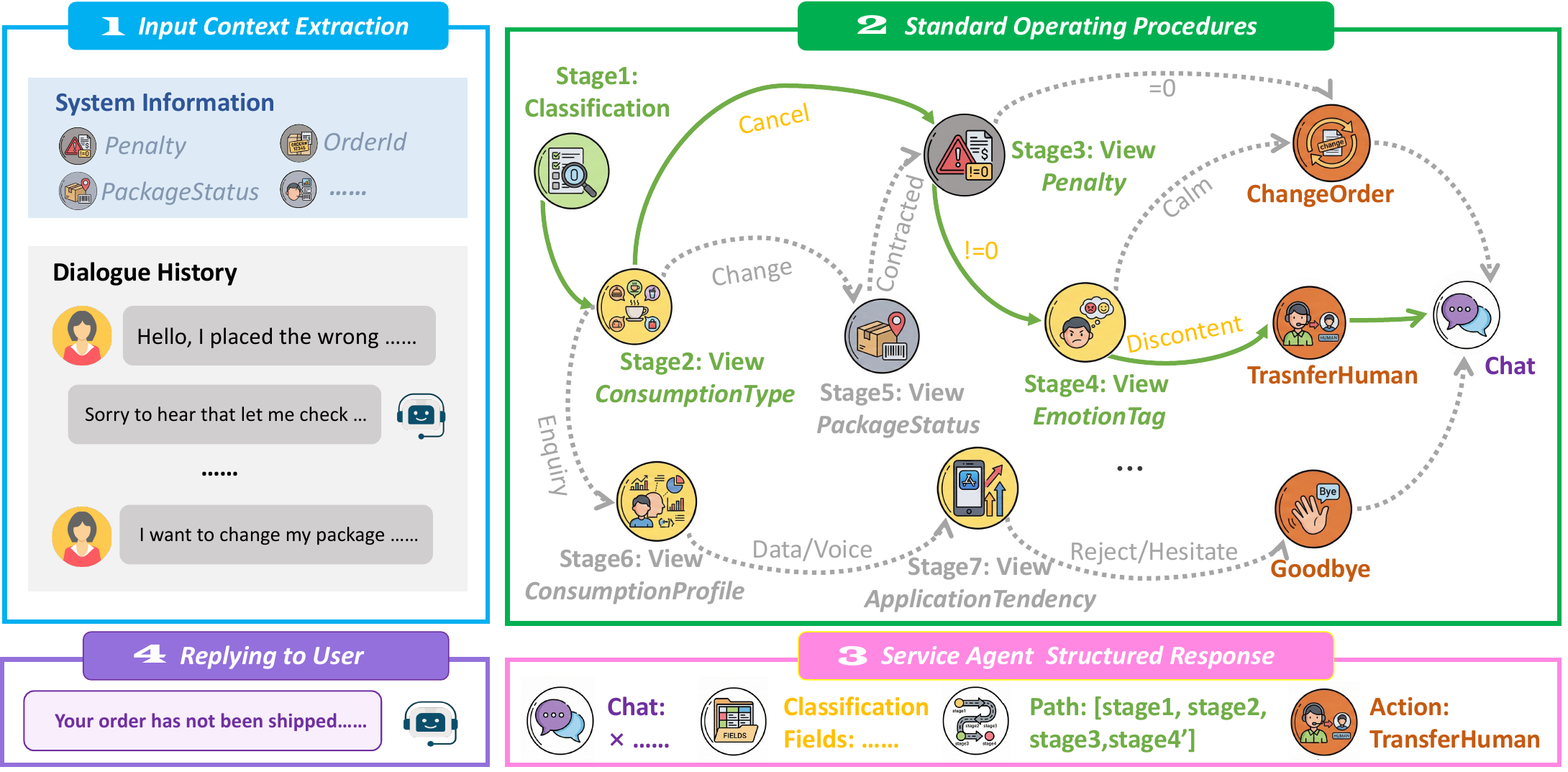} 
    \caption{Service Agent SOP Example (Telecom Scenario).
}
    \label{fig:intro}
\end{figure}

The rapid advancement of Large Language Models (LLMs) has significantly accelerated the automation process across various industrial sectors~\citep{bai2023qwen, touvron2023llama, wiggins2022opportunities}. Among these, the domain of intelligent customer service has emerged as a pioneer in adopting these technologies to enhance operational efficiency and user experience~\cite{cui2017superagent, zhao2024dialogbench, sun2024ecombench}. By leveraging the robust generative capabilities and context understanding of LLMs, enterprises are striving to transition from traditional, rigid rule-based systems to highly adaptive intelligent service agents~\cite{xi2025rise, wang2024survey} capable of handling complex interactions~\cite{yao2023react, shinn2023reflexion}, thereby addressing a broader range of customer needs while reducing labor costs.

Consequently, benchmarking models on their strict adherence to Standard Operating Procedures (SOPs) for correct workflow transitions, a process requiring the generation of structured responses that align with predefined logic as shown in Figure~\ref{fig:intro}, has become a priority for enterprises.
However, existing service benchmarks face three fundamental limitations. 
First, \textbf{insufficient evaluation dimensions}: Current benchmarks typically assess either task completion~\cite{qin2023toolllm, li2023api, liu2023agentbench} or dialogue quality~\cite{zheng2023judging, team2023vicuna} in isolation. Real-world scenarios, however, demand both strict \textit{logical compliance} with SOPs and \textit{appropriate communication skills}. 
This metric insufficiency causes evaluation bias and hinders precise error localization.
Second, \textbf{static interaction paradigms}: Relying on fixed datasets like scripts~\cite{budzianowski2018multiwoz, rastogi2020towards, zheng2023lmsyschat1m}, traditional methods fail to test error recovery or cover diverse user behaviors, ranging from cooperative inquiries to adversarial conflicts. Consequently, these static evaluations are incomplete and struggle to reflect real-world performance.
Finally, \textbf{limited scalability}: Benchmarks often depend on costly manual annotation~\cite{quan2020risawoz, li2025sopbench} and are frequently over-fitted to the SOPs of a single domain, making it prohibitively expensive to adapt to diverse, multi-branch business scenarios.

To address these challenges, we propose \textbf{SAGE} (\textbf{S}ervice \textbf{A}gent \textbf{G}raph-guided \textbf{E}valuation). SAGE integrates three core contributions: 
(1) \textbf{Dynamic Multi-Turn Dialogue Graph Modeling}, which formalizes SOPs into directed graphs (as shown in Figure~\ref{fig:intro}) to enable dynamic verification of logical compliance and ensure comprehensive path coverage;
(2) a \textbf{Multi-Agent multi-dimensional Evaluation} utilizes Judge Agents and a Rule Engine to analyze the interaction between User and Service Agents, generating deterministic ground truth for a rigorous, multi-dimensional assessment of logical compliance and chat quality; and (3) a \textbf{Scenario Extension Mechanism}, which enabled the rapid deployment of six industrial scenarios in our study, demonstrating its practical feasibility.

To validate SAGE, we evaluated 27 mainstream LLMs across 6 industrial scenarios. Our experiments reveal: (1) The gap between open-source and closed-source models is narrowing, with DeepSeek-V3.2 surpassing several GPT-4 class models; (2) A significant ``Execution Gap'' in complex scenarios, where high classification accuracy does not guarantee correct action execution, highlighting the challenge of procedural reasoning; (3) Performance degradation in multi-turn dialogues due to context fatigue; and (4) ``Empathy Resilience'' under high adversarial intensity, where models maintain polite conversational facades despite underlying logical failures.

In summary, our contributions are:
\begin{itemize}
    \item We propose \textbf{SAGE}, the first graph-guided multi-agent evaluation benchmark that transforms unstructured SOPs into directed graphs to enable automated, dual-axis assessment of logical compliance and conversational quality.
    \item We introduce dynamic graph modeling and an \textbf{Adversarial Intent Taxonomy}, bridging the gap between static testing and dynamic reality through diverse user behavior simulations.
    
    \item We uncover critical phenomena such as the {\textbf{``Execution Gap''}} and {\textbf{``Empathy Resilience''}}, providing granular diagnostics for agentic capabilities, through extensive experiments on 27 LLMs across 6 distinct scenarios.

    \item We design a modular \textbf{Extension Mechanism} for low-cost adaptation to new scenarios, which also facilitates the automated synthesis of large-scale dialogue datasets for customer service.

\end{itemize}

\section{Related Works}

\subsection{Evaluation for Large Language Models}
Large language model evaluation has evolved from basic instruction-following~\cite{zhou2023instruction} to complex multi-constraint scenarios. Early benchmarks assessed format compliance~\cite{xia2024fofo, tang2024struc}, while recent work evaluates real-world complexity through Multi-IF~\cite{he2024multi}, FollowBench~\cite{jiang2024followbench}, and InfoBench~\cite{qin2024infobench}. Guidebench~\cite{diao2025guidebench} introduces domain-specific conditions, and Collie~\cite{yao2023collie} systematically constructs constrained generation tasks.
Domain-specific evaluation spans business process management~\cite{berti2024pm, fahland2024well, kourani2025evaluating, fournier2024towards, rebmann2024evaluating, kourani2025leveraging, grohs2023large}, finance~\cite{xie2024finben, xie2023pixiu}, and procedural compliance. SOPBench~\cite{li2025sopbench} and SOP-Bench~\cite{nandi2025sopbench} evaluate tool-calling sequences but primarily focus on external manipulation rather than deep logical reasoning. 
Broader agent frameworks include AgentBench~\cite{liu2023agentbench}, WebArena~\cite{zhou2024webarena}, and WorkArena~\cite{drouin2024workarena}. Complementing mathematical tasks, textual logical reasoning, and reading comprehension are assessed through LogiQA~\cite{liu2020logiqa} for deductive reasoning, DROP~\cite{dua2019drop} for discrete reasoning over paragraphs, and HotpotQA~\cite{yang2018hotpotqa} for multi-hop reasoning, which are more closely aligned with the context understanding required in service scenarios.
Agent training advances include AGILE~\cite{zhang2024agile}, ReAct~\cite{yao2023react}, Reflexion~\cite{shinn2023reflexion}, Agent-Pro~\cite{zhang2024agentpro}, AgentTuning~\cite{zeng2023agenttuning}, Self-Refine~\cite{madaan2023selfrefine}, Self-Instruct~\cite{wang2023selfinstruct}, and human feedback training~\cite{ouyang2022training}.
Unlike prior benchmarks limited by static datasets and single-dimensional metrics, SAGE introduces a graph-guided multi-agent framework to dynamically evaluate both the logical compliance and conversational quality of service agents.

\subsection{Benchmarks for Service Dialogue Systems}
Service dialogue evaluation addresses customer-facing applications through DialogBench~\cite{zhao2024dialogbench} for human-like conversation, ECom-Bench~\cite{sun2024ecombench} for customer support resolution. Multi-turn complexity is captured by MG-ShopDial~\cite{li2024mgshopdial}, Wizard of Shopping~\cite{li2025wizard}, and Parrot~\cite{sun2024parrot}. Customer support dialogue is studied through evaluation frameworks~\cite{liu2024evaluating}, real-world conversation data~\cite{zheng2023lmsyschat1m}, and recommendation as instruction following~\cite{zhang2025recommendation}. Existing benchmarks often rely rigidly on scenario-specific Standard Operating Procedures (SOPs), limiting their adaptability. SAGE addresses this with a modular, intent-based extension mechanism that enables rapid, code-free adaptation to new domains.

\section{Methodology}
\begin{figure*}
    \centering
    \includegraphics[width=\linewidth]{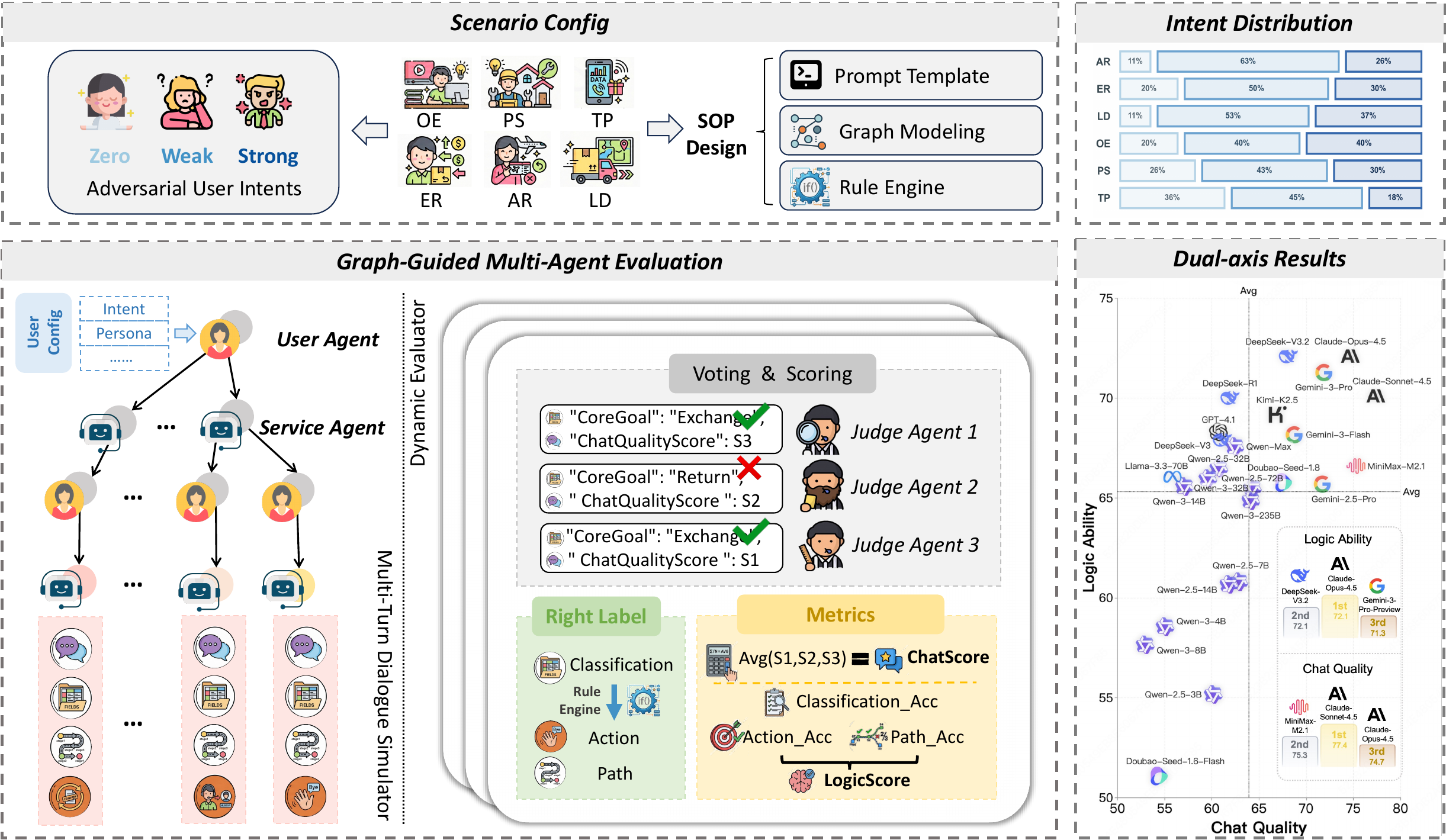}
    \caption{Overview of SAGE evaluation framework.}
    \label{fig:framework}
\end{figure*}

To address the complexity of evaluating customer service LLMs, we propose SAGE, a graph-guided multi-agent framework. As shown in Figure~\ref{fig:framework}, the workflow integrates three key stages: (1) \textbf{Dynamic Multi-Turn Dialogue Graph Modeling} formalizes SOPs into directed graphs to enable dynamic verification of logical compliance against correct paths while ensuring comprehensive coverage of all potential scenarios; (2) \textbf{Graph-Guided Multi-Agent Evaluation} rigorously assesses these trajectories; and (3) an \textbf{Scenario Extension Mechanism} that leverages both user intents and SOPs to enable rapid adaptation to arbitrary scenarios through modular configuration. We begin by detailing the graph formalization process.

\subsection{Dynamic Multi-Turn Dialogue Graph Modeling}

Aiming at a systematic evaluation of LLMs applied in service agent, we transform SOPs into computable graphs to underpin our framework, thereby allowing for automated logical verification and comprehensive scenario traversal.

\subsubsection{Procedure Graph Formalization}

Customer service Standard Operating Procedures (SOPs) typically exist as natural language documents containing numerous conditional branches as shown in Figure \ref{fig:intro}. Using such unstructured descriptions directly for automated evaluation is prone to ambiguity. Therefore, we formalize SOPs as directed graphs $G = (V, E)$. The node set $V$ consists of three types: (1) \textbf{Start/End Nodes}; (2) \textbf{Decision Nodes}, which branch the flow based on specific conditions; and (3) \textbf{Action Nodes}, representing concrete operations. The edges $E$ define the transition logic, where transitions are triggered by the values of specific \textbf{Classification Fields} $\mathcal{F}$ (e.g., Order Type) and \textbf{System Information} $\mathcal{S}$.
This graph structure serves as the backbone of SAGE. It functions both as the navigation map for the Service Agent and the evaluation standard for the Rule Engine. Crucially, \textbf{System Information} $\mathcal{S}$ is consistently propagated throughout the process: it is used to shape the User Agent's persona, utilized by the Service Agent for decision-making, and finally employed by the Rule Engine to generate process ground truth. This consistency ensures the fairness and determinism of the evaluation.

\subsubsection{User-Agent Multi-turn Interaction}

To overcome the limitations of traditional benchmarks, we generate dialogue trajectories through dynamic interaction between user agents and service agents.

\textbf{User Agent} generates user responses based on personas and dialogue context:
\begin{equation}
a_t = \text{UserAgent}(h_{<t}, s_t, \mathcal{S}, \mathcal{I}, \mathcal{P}).
\end{equation}
The user agent's role is to provide dynamic adversarial testing for the service agent. The generated message $a_t$ comprehensively considers five factors: (1) \textbf{dialogue history} $h_{<t}$, which references previous exchanges; specifically, when history is empty, a dedicated initialization function is triggered to generate the opening utterance to populate $h_{<t}$; (2) \textbf{agent state} $s_t$, utilized to align the user's response with the agent's current status, including handling opening protocols and detecting dialogue termination conditions; (3) \textbf{system information} $\mathcal{S}$, which grounds the user's knowledge in reality (e.g., knowing their own payment status) to prevent hallucinations and ensure logical consistency; and crucially, (4) \textbf{user intent} $\mathcal{I}$ and (5) \textbf{user persona} $\mathcal{P}$. As further detailed in Section~\ref{sec:taxonomy}, $\mathcal{I}$ determines the user's fundamental goal (e.g., refund vs. inquiry), while $\mathcal{P}$ characterizes the user persona (encompassing traits such as emotional state, communication style, and cooperativeness). Both are dynamically instantiated to simulate diverse adversarial intensities.

\textbf{Service Agent} is the evaluation target tasked with navigating the SOP graph to generate a structured response. The output at turn $t$, denoted as $b_t$, comprises four distinct components:
\begin{equation}
b_t = \text{ServiceAgent}(\mathcal{S}, h_{<t}, a_t, G) = \{p_t, \text{action}_t, \mathcal{F}_t, \text{chat}_t\}.
\end{equation}
Here, $G$ represents the SOP graph described by text. The components of $b_t$ are defined as follows: (1) $p_t$ denotes the \textbf{path}, representing the nodes the agent intends to traverse in the current turn; (2) $\text{action}_t$ is the executed \textbf{action}, selected from the allowable action set $\mathcal{A}$ defined by the current node; (3) $\mathcal{F}_t$ represents the \textbf{classification fields}, capturing the agent's categorical judgment of specific classification fields (e.g., user emotion type); and (4) $\text{chat}_t$ is the natural language \textbf{chat} response generated to interact with the user. The agent's core objective is to accurately identify the classification field $\mathcal{F}_t$, which dictates the subsequent transition path $p_t$ and the mandated action $\text{action}_t$ within graph $G$, ultimately conditioning the generation of the response $\text{chat}_t$.

Consequently, the User Agent acts as an environment, presenting a specific service scenario to the Service Agent. The Service Agent must then navigate multi-turn interactions to achieve specific goals while strictly adhering to the SOP. We evaluate these generated dialogue trajectories in the subsequent sections.

\subsection{Graph-Guided Multi-Agent Evaluation}
Upon the completion of dialogue trajectory generation, SAGE executes a systematic evaluation through a graph-guided, multi-agent collaborative mechanism. To ensure that all potential business scenarios and logical branches are rigorously tested, we enforce \textbf{Path Coverage} as the first component, verifying that the generated trajectories span the entire state space of the formalized SOP graph. Following this structural validation, SAGE performs a granular assessment of individual interactions. In this architecture, each node and its corresponding edge are evaluated through a hybrid mechanism: a \textbf{Judge Agent} extracts categorical labels and linguistic nuances from the dialogue, while a \textbf{Rule Engine} cross-validates these outputs against the graph-defined logic. This dual-layered approach allows SAGE to simultaneously assess \textit{Logical Compliance} and \textit{Chat Quality}, ensuring that agents are both procedurally correct and contextually appropriate.

\subsubsection{Path Coverage}
To guarantee full logic coverage, we predefine all valid trajectories $\mathcal{P}$ based on the SOP graph. This involves a two-stage process consisting of initial intent-balanced sampling followed by targeted supplementation for under-covered paths. For rare branches, we use an inverse configuration mechanism to deterministically synthesize the required user intents and system states. This ensures every logical path, including edge cases, is tested frequently, establishing a robust reference standard for the subsequent evaluation. With testing comprehensiveness secured, we next detail the methodology for evaluating each trajectory.

\subsubsection{Judge Agent}
The Judge Agent is designed to handle the semantic understanding of natural language interactions. For each turn $t$, it analyzes the system information $\mathcal{S}$, dialogue history $h_{<t}$, user message $a_t$, and the service agent's natural language response $\text{chat}_t$ (excluding internal reasoning steps). The judge outputs the classification ground truth $\mathcal{F}^*_t$ (e.g., user's emotion type and user's goal) and a conversational quality score $s_{\text{chat}}$:
\begin{equation}
(\mathcal{F}^*_t, s_{\text{chat}}) = \text{JudgeAgent}(\mathcal{S}, h_{<t}, a_t, \text{chat}_t).
\end{equation}
To ensure robustness against individual model bias, we employ an ensemble of three Judge Agents. We apply majority voting to determine the consensus ground truth $\mathcal{F}^*_t$ of classification fields, while the quality score $s_{\text{chat}}$ is derived from the average rating. The classification ground truth $\mathcal{F}^*_t$ serves as the input for the Rule Engine, while $s_{\text{chat}}$ directly quantifies the conversational quality.

\subsubsection{Rule Engine}
The Rule Engine functions as a deterministic generator for procedural logic ground truth. It receives the consensus classification ground truth $\mathcal{F}^*_t$ (derived from the Judge Agent), system information $\mathcal{S}$, and the SOP graph $G$ as inputs. By performing a deterministic search on the graph, it calculates the unique, theoretically correct execution path and action:
\begin{equation}
(p^*_t, \text{action}^*_t) = \text{RuleEngine}(\mathcal{F}^*_t, \mathcal{S}, G).
\end{equation}
Here, $p^*_t$ represents the reference path and $\text{action}^*_t$ denotes the reference action. These outputs serve as the rigid standard for evaluating the service agent's logical reasoning capabilities, specifically its path planning and action selection accuracy.

\subsubsection{Dual-Axis Evaluation Metrics}

With the ground truth established, we conduct a dual-axis evaluation comparing the service agent's output against these standards. This composite design facilitates the precise localization of model defects to specific granular dimensions.

\paragraph{Logical Compliance Evaluation.} We assess logical adherence across three dimensions, with weights $w_1=0.4, w_2=0.4, w_3=0.2$:
(1) Classification Accuracy measures the alignment between the agent’s predicted field $\mathcal{F}_t$ and the ground truth label determined by the Judge Agent. Specifically, it quantifies the agent's proficiency in correctly identifying the state-specific attributes that trigger graph transitions. A high Logic score indicates the agent correctly identifies intent and follows the SOP graph.
\begin{equation}
\text{Acc}_{\text{cls}} = |\mathcal{F}|^{-1} \sum_{f \in \mathcal{F}} \mathbb{I}[\mathcal{F}_t(f) = \mathcal{F}^*_t(f)].
\end{equation}
(2) Path Correctness measures the overlap between the agent's planned path and the Rule Engine's reference path:
\begin{equation}
\text{Sim}_{\text{path}} = \frac{|p_t \cap p^*_t|}{|p^*_t|}.
\end{equation}
(3) Action Correctness verifies if the agent's final executed action matches the Rule Engine's reference action:
\begin{equation}
\text{Acc}_{\text{action}} = \mathbb{I}[\text{action}_t = \text{action}^*_t].
\end{equation}

\paragraph{Chat Quality Evaluation.} This metric assesses the linguistic and interactive performance of the service agent. To ensure a multi-dimensional evaluation, the Judge Agents evaluate each response across five key dimensions: Linguistic Quality, Anthropomorphism, {Content Utility}, {User Satisfaction}, and {Instruction Compliance}.
The final score for this metric is derived through a two-step aggregation: first, for each individual judge, a weighted sum is calculated based on these five dimensions to reflect their relative importance; second, the scores from the three independent Judge Agents are averaged to mitigate subjective bias and ensure the reliability of the evaluation.

\begin{equation}
\text{Score}_{\text{quality}} = \frac{1}{3} \sum_{j=1}^{3} \left( \sum_{k=1}^{5} s_{j,k}\right).
\end{equation}
\paragraph{Overall Assessment Score.} The final overall score for a turn integrates both axes:
\begin{equation}
\text{Score}_{\text{overall}} = 0.8 \times \text{Score}_{\text{logic}} + 0.2 \times \text{Score}_{\text{quality}}.
\end{equation}
For multi-turn dialogues, we adopt a turn-level evaluation strategy (assessing turns 1, 5, 10, 15, and the final turn) to measure performance across different conversation depths.
This weighted integration prioritizes procedural rigor while accounting for user experience, facilitating the precise diagnosis of model defects across specific dimensions. Beyond robust evaluation, SAGE is engineered for scalability. We next detail how its modular design supports rapid adaptation to arbitrary scenarios.

\subsection{Scenario Extension Mechanism}
\label{sec:extension}

\textbf{Intent-based Adversarial Scenario Taxonomy.}
\label{sec:taxonomy}
To guarantee a comprehensive simulation of realistic user behaviors, we establish a taxonomy based on two critical dimensions: \textit{goal alignment} (the extent to which user demands match agent capabilities) and \textit{emotional state} (the intensity of user aggression or urgency):
\textbf{Zero-adversarial Intents} reflect cooperative interactions where user goals align with agent services (e.g., standard payment inquiries). Characterized by clear requests and friendly attitudes, these scenarios primarily test the agent's basic procedural execution. \textbf{Weak-adversarial Intents} introduce procedural friction or rational criticism. Here, users present complex constraints (e.g., unpaid bills) or ambiguous needs, requiring agents to resolve contextual conflicts without facing direct hostility. \textbf{Strong-adversarial Intents} represent high-stakes conflicts driven by emotional dissatisfaction or emergencies (e.g., safety hazards). Users employ aggressive strategies to demand immediate resolutions, rigorously testing the agent's negotiation, de-escalation, and risk control capabilities.

\textbf{Extension Mechanism.}
We implement this taxonomy via prompt engineering, integrating user intent $\mathcal{I}$ and persona $\mathcal{P}$ into the User Agent to generate diverse, high-fidelity scenarios (details in Appendix~\ref{app:scenario_extending}). Consequently, SAGE facilitates rapid scenario extension through a streamlined two-step process: formalizing the SOP graph and defining the user persona profile. This modular, ``fill-in-the-blank'' approach decouples scenario configuration from the core evaluation engine, significantly lowering the technical barrier for deployment. Beyond evaluation, this high-scalability architecture can be further extended to automated dialogue data synthesis, enabling the large-scale generation of high-quality training corpora for customer service LLMs.

\section{Experiments}
\label{sec:exp}

\begin{table*}[!t]
\centering
\caption{Main results across 6 scenarios (0-100 Scale). \textit{OA}: Overall Assessment Score; \textit{Logic}: Logical Compliance Score; \textit{Chat}: Conversational Quality Score. The superscripts indicate the ranking within Closed-Source and Open-Source groups, respectively. \textit{Format Error}: Percentage of outputs failing JSON parsing. \textit{Chat Length}: Average character count of the response field. }%主

\begin{tabular}{lr|ccc|cccccc|c|c}
\toprule
\multicolumn{2}{c|}{\textbf{Model}} & \multicolumn{3}{c|}{\textbf{AVG Score}} & \multicolumn{6}{c|}{\textbf{OA on 6 Scenarios}} & \multicolumn{1}{c|}{\textbf{Format}} &\multicolumn{1}{c}{\textbf{Chat}} \\
\cmidrule(lr){1-2} \cmidrule(lr){3-5} \cmidrule(lr){6-11} \cmidrule(lr){12-12} \cmidrule(lr){13-13}
\multicolumn{1}{l}{Name}& \multicolumn{1}{c|}{Params} & \textbf{OA} & \textbf{Logic} & \textbf{Chat} & ER & TP & PS & LD & AR & OE & \textbf{Error} & \textbf{Length}\\

\midrule

\rowcolor{gray!20}\multicolumn{13}{c}{\textbf{\textit{Closed-Source Large Language Models}}} \\ \midrule

Claude-Sonnet-4.5 & - & 71.56$^{2}$ & 70.10$^{3}$ & 77.38$^{1}$ & 71.21 & 75.03 & 75.07 & 67.30 & 66.52 & 74.24 & 0.75 \% & 93.42 \\
Claude-Opus-4.5 & - & 72.62$^{1}$ & 72.10$^{1}$ & 74.68$^{2}$ & 72.73 & 73.12 & 73.25 & 71.25 & 70.12 & 75.22 & 0.94 \% & 43.49 \\
GPT-4.1 & - & 66.79$^{6}$ & 68.31$^{4}$ & 60.70$^{8}$ & 70.32 & 66.90 & 65.46 & 62.69 & 63.68 & 71.69 & 0.04 \% & 18.59 \\
Gemini-2.5-Pro & - & 66.90$^{5}$ & 65.69$^{8}$ & 71.73$^{4}$ & 69.46 & 66.42 & 69.60 & 64.92 & 60.71 & 70.28 & 0.00 \% & 53.33 \\
Gemini-3-Pro-Preview & - & 71.40$^{3}$ & 71.27$^{2}$ & 71.89$^{3}$ & 70.99 & 69.09 & 73.65 & 68.05 & 64.45 & 82.15 & 0.00 \% & 48.60 \\
Gemini-3-Flash-Preview & - & 68.28$^{4}$ & 68.16$^{5}$ & 68.75$^{5}$ & 67.56 & 65.36 & 71.62 & 67.10 & 62.40 & 75.62 & 8.92 \% & 50.16 \\
Qwen-Max & 1T+ & 66.54$^{7}$ & 67.57$^{6}$ & 62.44$^{7}$ & 66.81 & 65.47 & 64.36 & 64.03 & 63.85 & 74.75 & 0.00 \% & 27.55 \\
Doubao-Seed-1.6-Flash & 230B & 51.75$^{9}$ & 51.08$^{9}$ & 54.43$^{9}$ & 54.18 & 51.59 & 47.34 & 46.98 & 47.37 & 63.04 & 0.49 \% & 18.83 \\
Doubao-Seed-1.8 & - & 66.14$^{8}$ & 65.78$^{7}$ & 67.62$^{6}$ & 70.25 & 65.89 & 69.03 & 64.01 & 62.19 & 65.49 & 0.04 \% & 26.79 \\
\midrule

\rowcolor{gray!20}\multicolumn{13}{c}{\textbf{\textit{Open-Source Large Language Models}}} \\ \midrule

Qwen2.5-3B-Instruct & 3B & 56.16$^{17}$ & 55.16$^{17}$ & 60.15$^{12}$ & 68.92 & 55.16 & 54.53 & 46.01 & 59.65 & 52.70 & 0.02 \% & 26.76 \\
Qwen2.5-7B-Instruct & 7B & 61.22$^{13}$ & 60.80$^{13}$ & 62.86$^{6}$ & 69.05 & 62.58 & 60.19 & 47.24 & 60.88 & 67.35 & 0.00 \% & 19.44 \\
Qwen2.5-14B-Instruct & 14B & 60.90$^{14}$ & 60.67$^{14}$ & 61.80$^{8}$ & 67.64 & 63.33 & 58.12 & 54.24 & 62.96 & 59.11 & 0.00 \% & 27.23 \\
Qwen2.5-32B-Instruct & 32B & 65.34$^{7}$ & 66.49$^{7}$ & 60.74$^{11}$ & 72.07 & 68.27 & 62.26 & 62.99 & 60.25 & 66.20 & 0.00 \% & 21.18 \\
Qwen2.5-72B-Instruct & 72B & 65.27$^{8}$ & 65.51$^{11}$ & 64.30$^{4}$ & 64.90 & 69.16 & 65.78 & 63.16 & 59.83 & 68.76 & 0.00 \% & 44.67 \\
Qwen3-4B & 4B & 57.86$^{15}$ & 58.56$^{15}$ & 55.05$^{16}$ & 64.71 & 62.72 & 56.57 & 44.36 & 56.47 & 62.31 & 0.04 \% & 24.12 \\
Qwen3-8B & 8B & 56.71$^{16}$ & 57.65$^{16}$ & 52.93$^{17}$ & 51.88 & 62.80 & 54.03 & 47.77 & 57.67 & 66.08 & 0.00 \% & 19.93 \\
Qwen3-14B & 14B & 63.88$^{12}$ & 65.57$^{10}$ & 57.11$^{14}$ & 63.79 & 68.01 & 64.81 & 56.13 & 63.78 & 66.74 & 0.43 \% & 25.54 \\
Qwen3-32B & 32B & 64.69$^{10}$ & 65.97$^{9}$ & 59.58$^{13}$ & 67.94 & 65.48 & 66.71 & 58.79 & 61.66 & 67.56 & 1.48 \% & 24.15 \\
Qwen3-235B-A22B & 235B & 64.72$^{9}$ & 64.87$^{12}$ & 64.13$^{5}$ & 66.73 & 64.97 & 61.47 & 64.77 & 62.37 & 68.02 & 1.34 \% & 32.85 \\
Deepseek-V3.2 & 671B & 71.29$^{1}$ & 72.08$^{1}$ & 68.11$^{2}$ & 71.51 & 70.61 & 70.71 & 70.87 & 65.84 & 78.17 & 0.00 \% & 27.48 \\
Deepseek-V3 & 671B & 66.54$^{6}$ & 67.90$^{5}$ & 61.11$^{10}$ & 69.62 & 71.60 & 67.57 & 60.08 & 61.58 & 68.77 & 0.00 \% & 18.50 \\
Deepseek-R1 & 671B & 68.39$^{3}$ & 70.00$^{2}$ & 61.94$^{7}$ & 70.16 & 70.39 & 70.58 & 65.75 & 64.03 & 69.40 & 0.97 \% & 38.08 \\
GLM-4.7 & 355B & 67.25$^{5}$ & 68.68$^{4}$ & 61.52$^{9}$ & 68.14 & 66.02 & 69.64 & 66.50 & 60.82 & 72.36 & 1.46 \% & 17.58 \\
Kimi-K2.5 & 1T & 68.71$^{2}$ & 69.16$^{3}$ & 66.93$^{3}$ & 69.69 & 69.41 & 71.84 & 65.74 & 64.45 & 71.15 & 0.52 \% & 34.17 \\
MiniMax-M2.1 & 229B & 68.33$^{4}$ & 66.59$^{6}$ & 75.29$^{1}$ & 68.33 & 70.12 & 64.07 & 68.54 & 66.37 & 72.55 & 0.82 \% & 61.27 \\
Llama-3.1-8B-Instruct & 8B & 37.62$^{18}$ & 37.36$^{18}$ & 38.64$^{18}$ & 49.58 & 55.83 & 37.58 & 28.26 & 38.14 & 16.30 & 44.33 \% & 44.70 \\
Llama-3.3-70B-Instruct & 70B & 64.02$^{11}$ & 66.06$^{8}$ & 55.85$^{15}$ & 68.60 & 64.74 & 63.87 & 60.27 & 62.93 & 63.71 & 0.03 \% & 13.79 \\
\bottomrule
\end{tabular}
\label{tab:main_results}
\end{table*}

\subsection{LLM Configuration}

To ensure a comprehensive assessment of the current landscape, we select 27 representative Large Language Models (LLMs), categorized into closed-source and open-source families, covering a wide spectrum of parameter scales and architectures.

\textbf{Closed-Source Models.} We evaluate state-of-the-art proprietary systems accessed via official APIs. This includes the \textbf{Claude} series (Sonnet-4.5, Opus-4.5), known for strong reasoning capabilities; the \textbf{GPT} series (GPT-4.1)~\cite{achiam2023gpt}, serving as a standard baseline; and the \textbf{Gemini} series (2.5-Pro, 3-Pro/Flash)~\cite{team2023gemini}, representing multimodal-native architectures. We also include leading Chinese proprietary models such as \textbf{Qwen-Max}~\cite{bai2023qwen} and the \textbf{Doubao} series, which are widely optimized for Chinese application scenarios.

\textbf{Open-Source Models.} We cover models ranging from lightweight (3B) to massive scale (1T) to analyze the impact of model size. Our selection features the \textbf{Qwen2.5} and \textbf{Qwen3} families~\cite{bai2023qwen}, which provide a granular range of sizes (3B to 235B); the \textbf{DeepSeek} series (V3, V3.2, R1)~\cite{liu2024deepseek}, representing advanced Mixture-of-Experts (MoE) architectures; and the \textbf{Llama-3} series~\cite{dubey2024llama}. Additionally, we evaluate high-performing models from other providers, including \textbf{GLM-4.7}~\cite{zeng2022glm}, \textbf{Kimi-K2.5}~\cite{team2025kimi}, and \textbf{MiniMax-M2.1}~\cite{chen2025minimax}.

All open-source models are deployed locally using vLLM~\cite{kwon2023efficient} to ensure consistent inference efficiency, while closed-source models are evaluated using their respective stable API endpoints.

\subsection{Scenario Configuration}
\label{app:extension}
To evaluate the generalization capability of service agents across diverse industrial domains, we constructed six distinct customer service scenarios (detailed in Appendix~\ref{app:six_scenarios}). These scenarios range from standardized inquiries to complex, high-stakes disputes, covering varying levels of SOP complexity.

\begin{itemize}[left=0pt]
    \item \textbf{Ecommerce Refund (ER):} The most complex scenario, featuring a deep decision tree based on product status and credit levels. Agents must navigate multi-branch logic and negotiate terms with varied user temperaments.

    \item \textbf{Logistics Delivery (LD):} Centered on supply chain exceptions (e.g., lost or delayed parcels), requiring proficiency in status tracking and insurance claim processing.

    \item \textbf{Telecom Package (TP):} A standardized scenario focusing on linear SOP execution for billing and plan upgrades, evaluating instruction-following and upselling protocols.

    \item \textbf{Property Service (PS):} Emphasizes community coordination and emotional management (e.g., repair schedules or noise complaints) within offline service contexts.

    \item \textbf{Airline Refund (AR):} A high-complexity scenario governed by rigid, time-sensitive policies. It tests the agent's precision in calculating dynamic cancellation fees and de-escalating passenger anxiety.

    \item \textbf{Online Education (OE):} Focuses on long-term contract disputes and rigorous risk control. Agents must identify potential malicious refunders and strictly adhere to intricate refund formulas.

\end{itemize}

These scenarios collectively cover the spectrum from simple procedural execution to complex adversarial negotiation, ensuring a robust assessment of agentic capabilities.

\begin{figure*}[!t]
    \centering
    % --- 第一张图 ---
    % 宽度设为 0.32\textwidth，给间隔留出空间
    \begin{minipage}[t]{0.32\textwidth}
        \centering
        % 图片宽度设为 minipage 的宽度
        \includegraphics[width=\linewidth]{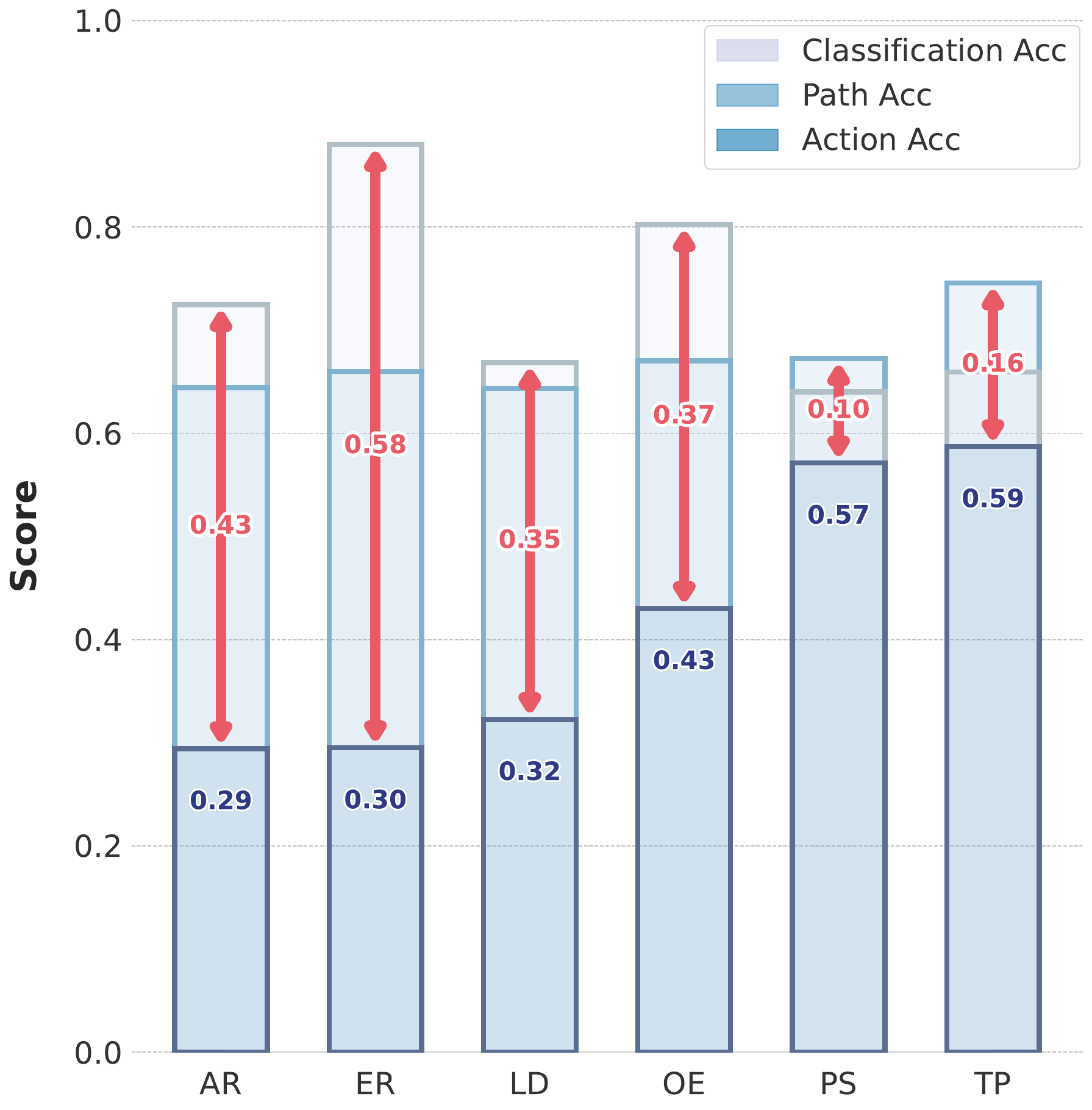}
        \caption{Logic performance gap analysis across six scenarios. 
        }
        \label{fig:acc_gap}
    \end{minipage}
    \hfill % 弹性空格，将图片推向两侧
    % --- 第二张图 ---
    \begin{minipage}[t]{0.32\textwidth}
        \centering
        \includegraphics[width=\linewidth]{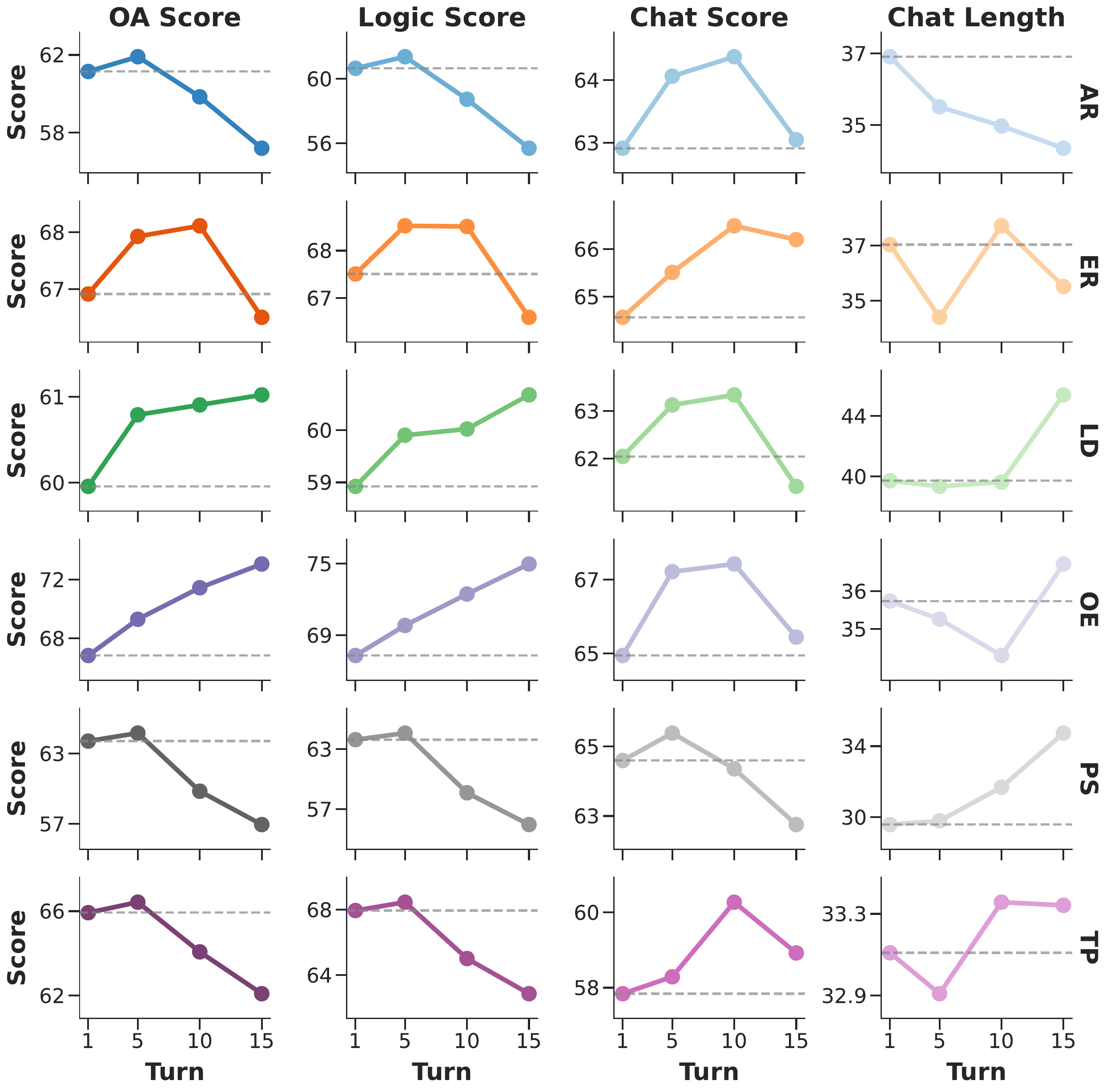}
        \caption{Performance evolution across dialogue turns in six scenarios. }
        \label{fig:turn_analysis}
    \end{minipage}
    \hfill % 弹性空格
    % --- 第三张图 ---
    \begin{minipage}[t]{0.32\textwidth}
    %     \centering
    % \includegraphics[width=1\linewidth]{fig/1_heatmap.png}
    % \caption{Enter Caption}}
    % \label{fig:placeholder}
        \centering
    \includegraphics[width=\linewidth]{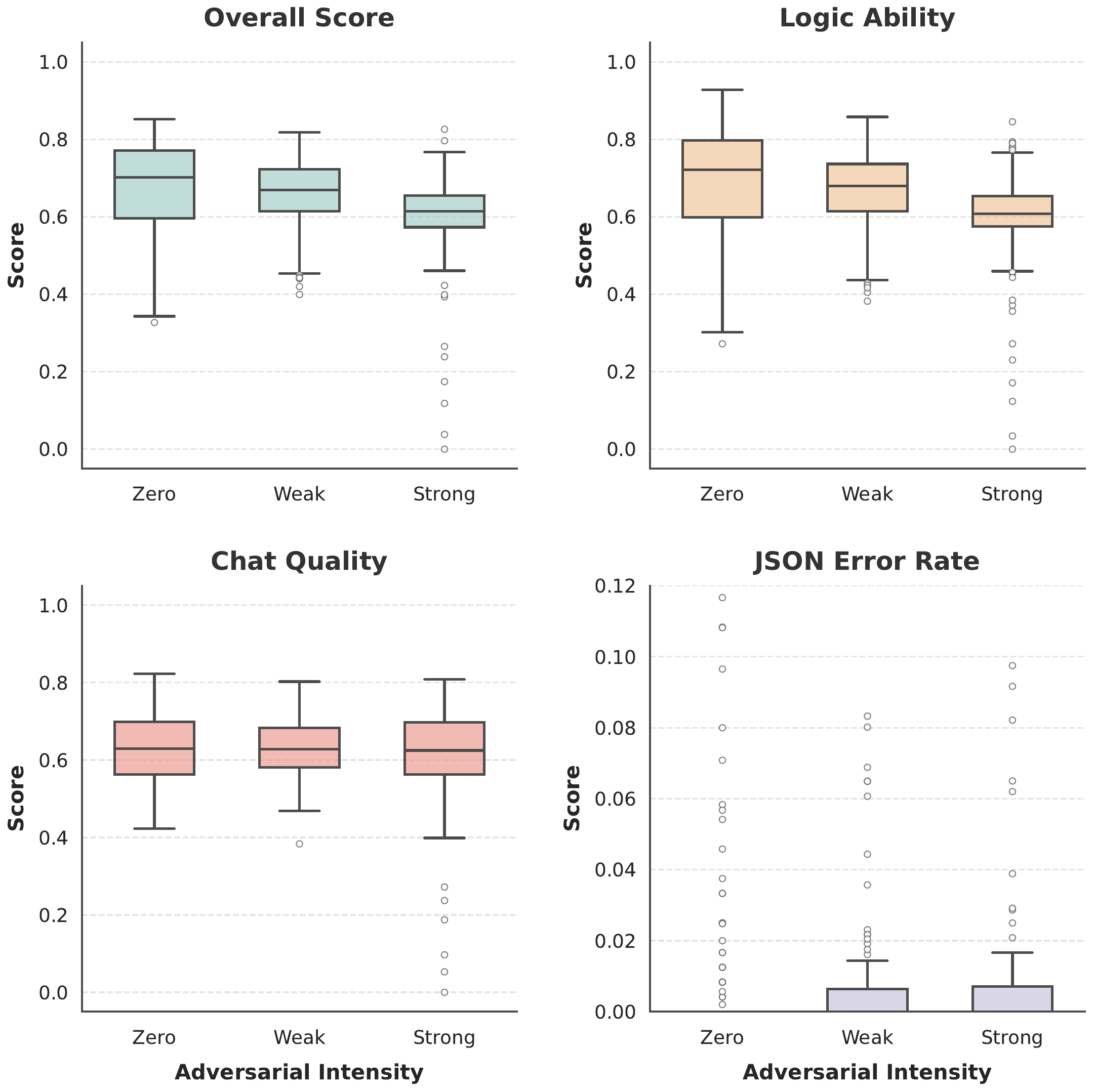}
    \caption{Performance distribution across 3-level Adversarial intensities. }
    \label{fig:adversarial_impact}
    \end{minipage}
\end{figure*}

\subsection{Model Performance Across Scenarios}
In addition to the primary metrics (Logical Compliance Score, Chat Quality Score and Overall Assessment Score), we report two auxiliary indicators to provide a more nuanced analysis of model behavior: Format Error Rate and Average Chat Length. The former quantifies the frequency of JSON parsing failures to reflect the model's instruction-following stability, while the latter measures response verbosity to identify potential issues with redundant generation or lack of conciseness. We evaluated 27 mainstream Large Language Models (LLMs), covering both closed-source (e.g., GPT-4, Claude-3.5) and open-source (e.g., Qwen2.5, Llama-3) families. Table~\ref{tab:main_results} presents the comprehensive performance across six diverse customer service scenarios.

\textbf{Superiority of Closed-Source Models and the Rising Open-Source Challengers.} Closed-source models continue to define the performance frontier, with Claude-Opus-4.5 securing the highest Overall Assessment (OA) score ($72.62^{1}$), underpinned by its top-tier logical reasoning ($72.10^{1}$) and impressive chat quality score ($74.68^{2}$). However, the performance gap between proprietary and open-weight models is remarkably narrow. DeepSeek-V3.2 ($671$B) emerges as a formidable competitor, achieving an OA of $71.29^{1}$ among open-source models—surpassing established closed-source giants like GPT-4.1 and Gemini-3-Pro-Preview. This indicates that state-of-the-art open-source architectures have reached a level of maturity capable of handling complex, graph-guided service logic previously reserved for proprietary systems.

\textbf{Decoupling Logical Compliance and Conversational Quality.} Our results reveal a nuanced trade-off between procedural rigor and linguistic flair. While Claude-Opus-4.5 leads in logic, its sibling Claude-Sonnet-4.5 dominates the Chat Quality category ($77.38^{1}$), suggesting a more empathetic persona. A notable outlier is MiniMax-M2.1, which, despite a moderate Logic score ($66.59^{6}$), achieves a stellar Chat score ($75.29^{1}$), the highest among open-source models. This suggests that certain models are specifically optimized for human-like interaction, which can occasionally compensate for minor procedural deviations in terms of overall user perception. In contrast, DeepSeek-R1 and GPT-4.1 lean heavily toward logic-first strategies, often at the expense of conversational warmth.

\textbf{The Dynamics of Response Verbosity and Stability.} We observe that response length is a significant indicator of service quality, but only up to a threshold. Top-tier performers like Claude-Sonnet-4.5 ($93.42$ chars) and MiniMax-M2.1 ($61.27$ chars) tend to provide more elaborate guidance and empathetic de-escalation, which are crucial in high-complexity scenarios like ER and AR. Conversely, the failure of smaller models is often catastrophic rather than gradual; for instance, Llama-3.1-8B suffers from an extreme Format Error rate ($44.33\%$), failing even the basic instruction-following required to output a valid JSON. Interestingly, Gemini-3-Flash-Preview maintains a competitive OA despite a higher-than-average error rate ($8.92\%$), indicating that when it does follow the format, its reasoning is remarkably efficient.

\subsection{Cross-Scenario Difficulty Analysis}

SAGE provides a standardized framework to quantify scenario complexity. As illustrated in Figure~\ref{fig:acc_gap}, we analyze three subdimension of logic performance across six domains. We define scenario difficulty by the ``Execution Gap''—the disparity between the nested bars, which reveal two distinct failure modes in LLM reasoning. First, in high-complexity scenarios such as \textbf{ER}, \textbf{AR}, \textbf{LD}, and \textbf{OE}, we observe a significant gap between Classification\_Acc (Light Gray) and Action\_Acc (Dark Blue). This disparity highlights a Logic Deduction Barrier: even when models correctly identify the classification fields ($\mathcal{F}_t$), they frequently fail to derive the correct subsequent action. This suggests that for complex SOPs, correct semantic classification does not naturally guarantee successful procedural execution due to intricate conditional dependencies.

Second, in more standardized scenarios like \textbf{PS} and \textbf{TP}, an interesting phenomenon occurs where Path\_Acc (Light Blue) exceeds Classification\_Acc (Light Gray). This stems from the metric's definition: Path Accuracy is calculated as the intersection length of the predicted and ground-truth paths divided by the total ground-truth length. Because this metric accounts for the successful traversal of early, simpler steps, models can achieve relatively high path scores by following partial correct fields, even if they stumble on specific complex classifications. This confirms that Path\_Acc serves as a more lenient dimension, whereas the gap between classification and action acts as a more rigorous stress test for deep procedural reasoning.

\subsection{Multi-turn Robustness Analysis}
\label{sec:multi-turn}
To evaluate model stability over extended interactions, we analyze performance variations across different dialogue depths (Turn 1, 5, 10, 15). Figure~\ref{fig:turn_analysis} illustrates the evolution of OA, Logic, and Chat scores across six scenarios.
A distinct performance trend \textbf{``Inverted-U'' Trajectory} is observed across most scenarios (e.g., AR, ER, PS, TP): scores typically peak at Turn 5 and decline significantly by Turn 15. This phenomenon can be attributed to two phases. \textbf{(1) Information Gain Phase (Turn 1 $\to$ 5):} Performance generally improves from the first turn to the fifth (e.g., ER OA rises from $\sim$67 to $\sim$68). In the initial turn (Cold Start), agents lack sufficient context. By Turn 5, through multi-turn interaction, agents gather critical user information (e.g., order IDs, specific complaints), enabling more accurate intent classification and SOP navigation. \textbf{(2) Context Fatigue Phase (Turn 10 $\to$ 15)}: As the dialogue extends beyond Turn 10, performance exhibits a marked decline (e.g., PS OA drops from $\sim$63 to $\sim$57). This degradation highlights the limitations of current LLMs in handling long-context dependencies. The accumulation of historical information introduces noise, leading to the ``Lost in the Middle'' phenomenon where models hallucinate or lose track of the current state within the SOP graph.

\subsection{Impact of Adversarial Intensity}
\label{sec:ad_intens}
To validate the effectiveness of our \textbf{Adversarial Scenario Taxonomy}, we analyze model performance across three intensity levels: Zero-, Weak-, and Strong-Adversarial. Figure~\ref{fig:adversarial_impact} presents the distribution of four key metrics—Overall Score (OA), Logic Score, Chat Quality, and Format Error Rate—via box plots.

\textbf{Logic Degradation and Error Escalation.}
As adversarial intensity increases, we observe a consistent decline in \textbf{Logical Compliance} (Logic Score). This trend confirms that adversarial user behaviors (e.g., concealing information, emotional aggression) successfully introduce logical friction, making it harder for agents to adhere to SOPs. Concurrently, the \textbf{Format Error Rate} rises significantly in Strong-Adversarial scenarios. This suggests that under high cognitive load or emotional pressure, models are more prone to instruction drift, failing to maintain the structured JSON output format required by the system.

\textbf{The Stability in Chat Quality.}
The distribution of Chat Quality scores remains stable across varying adversarial intensities. This ``Empathy Resilience'' reveals a decoupling between dialogue and logic: models maintain polite, de-escalating facades even when increased user aggression impairs their logical compliance. Such consistency underscores the need for dual-axis evaluation to distinguish surface fluency from procedural robustness.

\textbf{Increased Discriminative Power.}
Crucially, the score distribution becomes significantly more dispersed (larger interquartile range and more outliers) as intensity increases. In Zero-Adversarial settings, most models perform comparably well. However, Strong-Adversarial scenarios widen the gap between top-tier and lower-tier models. This proves that high-intensity scenarios serve as a more effective filter for distinguishing robust agentic capabilities.

\subsection{Correlation Analysis of Sub-Metrics}
\label{sec:sub_metric}

\begin{figure}[!t]
    \centering
    \includegraphics[width=0.9\linewidth]{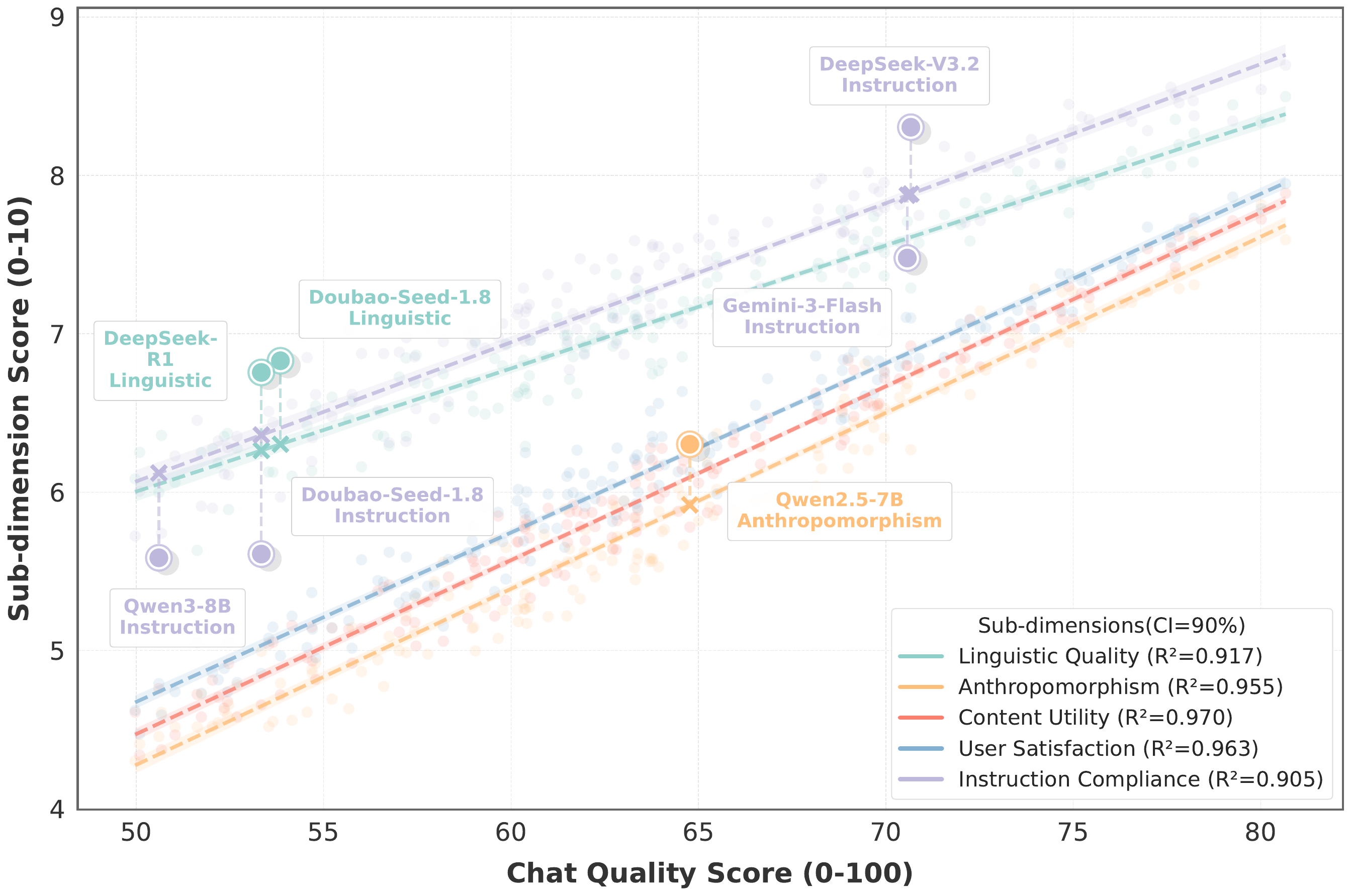}
\caption{Correlation analysis between the Chat Quality Score and its five sub-dimensions. The deviations from the regression lines (e.g., {Doubao-Seed-1.8}'s high Linguistic vs. low Instruction scores) highlight specific model characteristics.} 
    
    \label{fig:correlation}
\end{figure}

To validate the internal consistency of our evaluation framework and diagnose fine-grained model capabilities, we analyze the correlation between the aggregated \textbf{Chat Quality Score} (X-axis, scaled to 0-1) and its five constituent \textbf{Sub-dimension Scores} (Y-axis, scaled 0-10). As illustrated in Figure~\ref{fig:correlation}, the regression analysis yields several key insights.

\textbf{High Metric Consistency.} 
We observe an extremely strong linear correlation across all five dimensions—Linguistic Quality, Anthropomorphism, Content Utility, User Satisfaction, and Instruction Compliance—with coefficient of determination ($R^2$) values consistently exceeding \textbf{0.9}. This statistical coherence confirms the validity of our metric design. In complex multi-agent evaluation, a high $R^2$ indicates that each dimension provides a consistent and reliable contribution to the aggregated Chat Quality Score. If a specific dimension (e.g., User Satisfaction) exhibited a non-linear correlation—such as an S-shaped curve—or significant noise, it would suggest potential flaws in the scoring rubrics or latent hallucinations and inconsistencies in the LLM-as-judge. The absence of such anomalies in our results confirms that the SAGE evaluation framework effectively encapsulates the multi-faceted nature of conversational ability, maintaining high interpretability and minimal bias across all predefined dimensions.

\textbf{Capability Profiling via Variance.}
While the general trend is linear, the \textbf{vertical variance} (residuals) from the regression lines serves as a fingerprint for specific model strengths and weaknesses. A point significantly deviating from the mean regression line indicates that a model's capability in that specific dimension is disproportionate to its overall performance. Points located significantly \textit{above} the green regression line (Linguistic Quality) represent models with exceptional fluency and expression. Specifically, \textbf{Doubao-Seed-1.8} and \textbf{DeepSeek-R1} exhibit positive residuals in this dimension, indicating that their linguistic generation capabilities are superior to the average level expected for their score range. Points falling \textit{below} the purple regression line (Instruction Compliance) reveal deficits in following specific constraints. Notably, despite reasonable overall scores, \textbf{Doubao-Seed-1.8} and \textbf{Qwen3-8B} appear significantly below the trend line for Instruction Compliance. This suggests a capability imbalance: these models are highly articulate (high Linguistic score) but prone to ignoring specific formatting or constraint instructions (low Instruction score). This granular analysis proves that SAGE can diagnose subtle trade-offs in model alignment, distinguishing between models that are merely chatty and those that are strictly compliant.

\section{Conclusion}
In this paper, we addressed the limitations of existing customer service benchmarks—specifically their single-dimensional metrics, static interactions, and limited scalability—by proposing SAGE, a graph-guided multi-agent evaluation benchmark. By formalizing Standard Operating Procedures (SOPs) into dynamic graph structures and incorporating adversarial user simulation with a dual-axis evaluation mechanism, SAGE achieves a comprehensive assessment of service agents in complex business logic and multi-turn interactions.
Our extensive experiments not only validated SAGE's effectiveness in distinguishing model capabilities but also revealed critical insights, such as the ``Inverted-U'' performance degradation in long contexts and logical fragility under high adversarial intensity. Furthermore, SAGE's modular design ensures rapid extensibility to diverse vertical domains via simple configuration. We envision SAGE as a standardized metric tool to drive the evolution of intelligent customer service from simple chatbots to expert-level agents. Future work will explore more complex graph structures (e.g., nested subgraphs) and multimodal interaction scenarios.

\bibliographystyle{ACM-Reference-Format}
\bibliography{sample-base}

\appendix

\section{Experiment Results Supplementary}

This appendix provides supplementary data substantiating our main findings. We first validate our multi-agent ensemble via a Single-Judge Bias ablation study. Next, we present granular turn-level analysis to detail multi-turn robustness, followed by a breakdown of performance shifts under varying adversarial intensities. Finally, we decompose Logic and Chat scores into sub-metrics, quantifying the ``Execution Gap'' and capability imbalances.

\begin{figure*}[htbp]
  \centering
  % 第一张图
  \begin{minipage}[t]{0.32\textwidth}
    \centering
    \includegraphics[width=\linewidth]{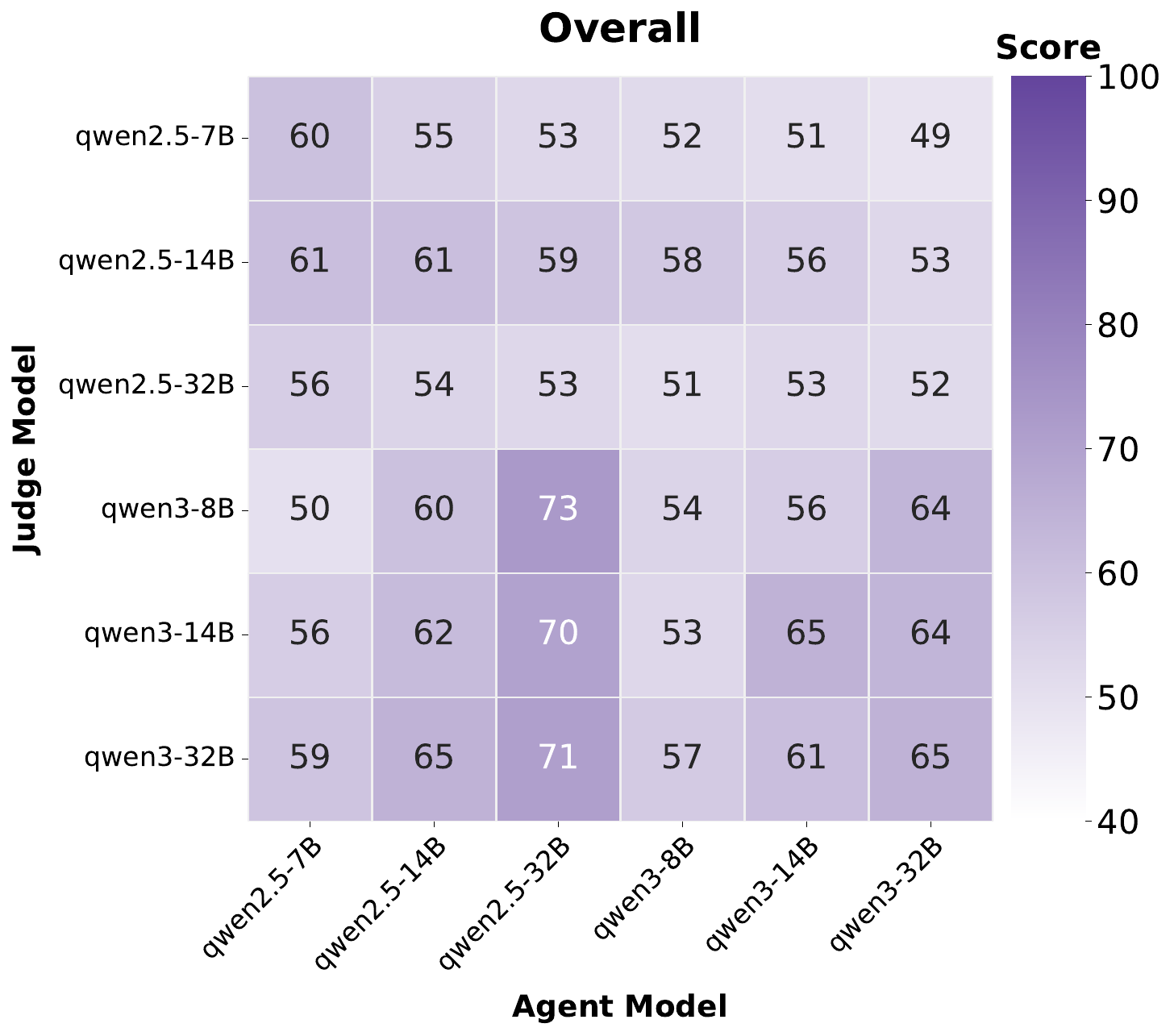} % 替换为实际文件名
    \caption{Overall Average Score (OA) Heatmap. Rows represent Judge models, columns represent Agent models.}
    \label{fig:heatmap_oa}
  \end{minipage}
  \hfill % 填充空白，使间距均匀
  % 第二张图
  \begin{minipage}[t]{0.32\textwidth}
    \centering
    \includegraphics[width=\linewidth]{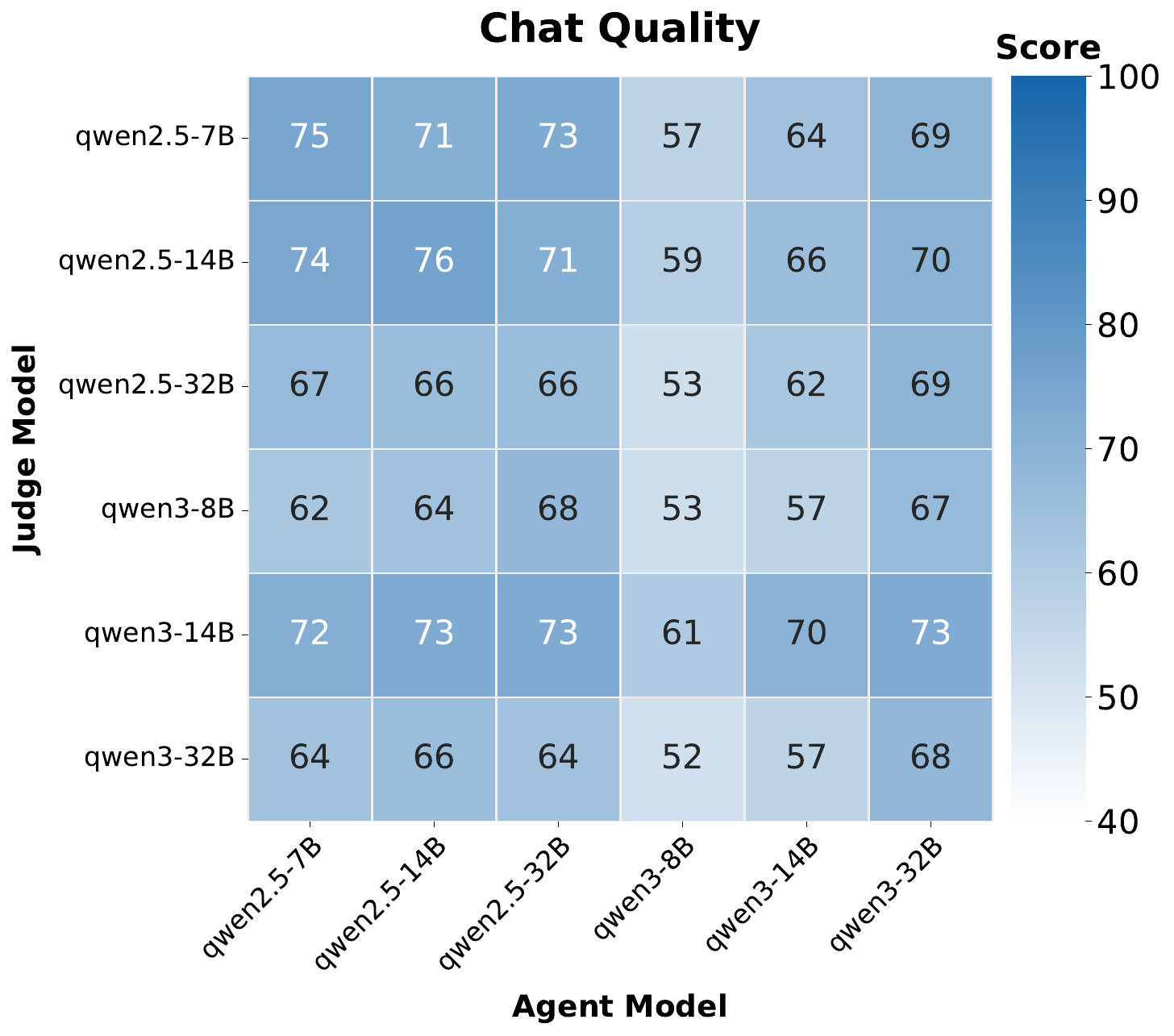}
    \caption{Chat Quality Heatmap. Darker diagonals indicate self-preference bias.}
    \label{fig:heatmap_chat}
  \end{minipage}
  \hfill
  % 第三张图
  \begin{minipage}[t]{0.32\textwidth}
    \centering
    \includegraphics[width=\linewidth]{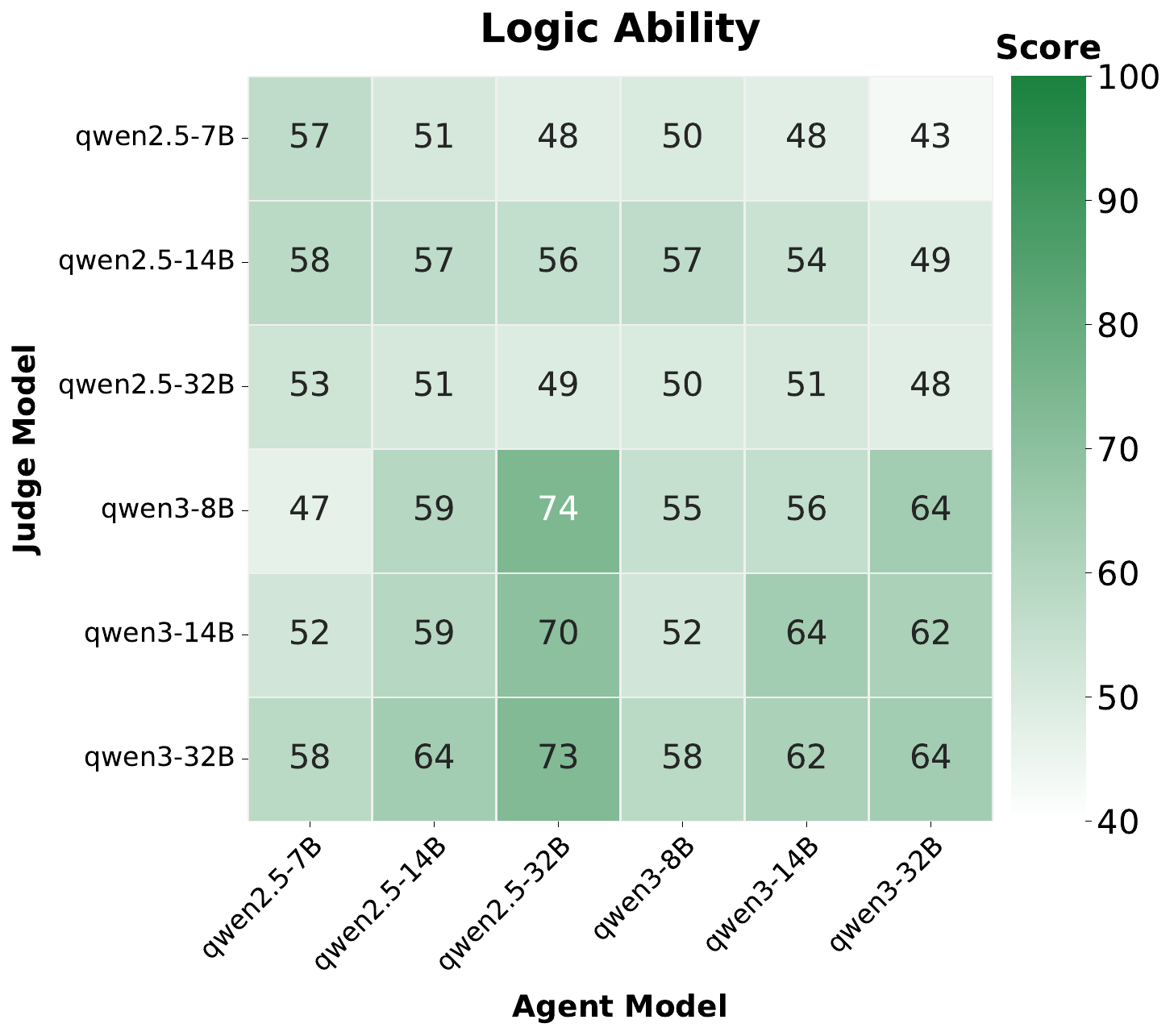}
    \caption{Logic Ability Heatmap. Horizontal variances indicate differing judge strictness.}
    \label{fig:heatmap_logic}
  \end{minipage}

\label{fig:judge_bias_heatmap}
\end{figure*}

\subsection{Impact of Single-Judge Bias}
\label{app:judge_bias}
To validate the robustness of our evaluation mechanism, we conducted an ablation study to investigate the bias inherent in using a single Large Language Model (LLM) as a judge. Specifically, we selected a subset of models (from the Qwen2.5 and Qwen3 families) to act as both "Service Agents" and "Judge Agents" in a round-robin evaluation setup.

The results are visualized in three heatmaps: \textbf{Overall Average Score (Figure~\ref{fig:heatmap_oa})}, \textbf{Chat Quality (Figure~\ref{fig:heatmap_chat})}, and \textbf{Logic Ability (Figure~\ref{fig:heatmap_logic})}. In these plots, the Y-axis represents the \textit{Judge Model} and the X-axis represents the \textit{Evaluated Agent}. This experiment reveals two critical limitations of single-judge frameworks:

\textbf{1. Egocentric Bias (Self-Preference).}
A distinct "diagonal dominance" is observable, particularly in \textbf{Figure~\ref{fig:heatmap_chat} (Chat Quality)}. Models tend to assign higher scores to their own outputs (or outputs from the same model family) compared to external evaluators. For instance, the diagonal cells in Figure~\ref{fig:heatmap_chat} often exhibit deeper colors than the off-diagonal cells in the same column, indicating that a model favors response styles similar to its own training distribution.

\textbf{2. Systematic Scoring Bias.}
Significant horizontal variations exist across all three heatmaps, especially in \textbf{Figure~\ref{fig:heatmap_logic} (Logic Ability)}. This indicates that different judges possess different strictness standards. Some judges (represented by rows with consistently lighter colors) act as "strict graders," systematically assigning lower scores across all agents, while others (darker rows) are more "lenient." For example, \textit{Qwen3-8B} as a judge might exhibit a different scoring distribution compared to \textit{Qwen2.5-32B}.

\textbf{Justification for Multi-Agent Ensemble.}
These findings demonstrate that relying on a single judge introduces significant variance and bias, where the evaluation outcome is heavily contingent on the specific evaluator's preferences rather than the intrinsic quality of the service agent. To mitigate this, SAGE employs an \textbf{Ensemble of Three Judge Agents} (using majority voting for classification and average scoring for quality). This collaborative approach effectively smooths out individual model biases, cancels out extreme scoring tendencies, and ensures a more objective and robust evaluation standard, as reflected in our main experimental results.

\subsection{Supplementary Analysis of Turn}
\label{app:turn}
This chapter supplements the content of Section~\ref{sec:multi-turn}.
Table~\ref{tab:turn_analysis} provides a granular view of model performance evolution across dialogue turns (T1, T5, T10, T15). Consistent with the ``Inverted-U'' trajectory discussed in the main text, most scenarios (e.g., AirlineRefund, PropertyService) exhibit a performance peak around Turn 5, followed by a decline at Turn 15.
\begin{itemize}
    \item \textbf{Context Accumulation (T1 to T5):} The initial improvement in Logic Score (e.g., TelecomPackage: $68.0$ to $68.5$) indicates that models effectively gather user information in early turns to clarify intent.
    \item \textbf{Context Fatigue (T10 to T15):} The subsequent drop (e.g., PropertyService Logic: $58.6$ to $55.5$) highlights the difficulty of maintaining logical consistency over long contexts.
    \item \textbf{Chat Stability:} Interestingly, Chat Scores often remain stable or decline less than Logic Scores (e.g., EcommerceRefund Chat: $66.5$ to $66.2$), suggesting that models maintain linguistic fluency even when their procedural reasoning falters.
\end{itemize}

\begin{table*}[htbp]
\centering
\caption{Turn-by-Turn Performance Analysis by Scenario. (0-100 Scale).}
\label{tab:turn_analysis}

\begin{tabular}{l|cccc|cccc|cccc|cccc}
\toprule
& \multicolumn{4}{c|}{\textbf{OA Score}} & \multicolumn{4}{c|}{\textbf{Logic Score}} & \multicolumn{4}{c|}{\textbf{Chat Score}} & \multicolumn{4}{c}{\textbf{Chat Length}} \\
\cmidrule(lr){2-5} \cmidrule(lr){6-9} \cmidrule(lr){10-13} \cmidrule(lr){14-17}
\textbf{Scenario} & T1 & T5 & T10 & T15 & T1 & T5 & T10 & T15 & T1 & T5 & T10 & T15 & T1 & T5 & T10 & T15 \\
\midrule

AirlineRefund & 61.1 & 61.9 & 59.8 & 57.2 & 60.7 & 61.4 & 58.7 & 55.7 & 63.1 & 64.1 & 64.3 & 63.2 & 37 & 36 & 35 & 34 \\
EcommerceRefund & 66.9 & 67.9 & 68.1 & 66.5 & 67.5 & 68.5 & 68.5 & 66.6 & 64.6 & 65.5 & 66.5 & 66.2 & 37 & 34 & 38 & 36 \\
LogisticsDelivery & 59.5 & 60.5 & 60.7 & 60.8 & 58.9 & 59.9 & 60.0 & 60.7 & 62.0 & 63.1 & 63.3 & 61.4 & 40 & 39 & 40 & 45 \\
OnlineEducation & 67.0 & 69.4 & 71.5 & 73.1 & 67.5 & 69.9 & 72.6 & 75.0 & 65.1 & 67.3 & 67.3 & 65.5 & 35 & 35 & 34 & 36 \\
PropertyService & 64.1 & 64.7 & 59.8 & 56.9 & 63.9 & 64.6 & 58.6 & 55.5 & 64.6 & 65.4 & 64.4 & 62.8 & 30 & 30 & 32 & 35 \\
TelecomPackage & 65.9 & 66.4 & 64.1 & 62.1 & 68.0 & 68.5 & 65.0 & 62.9 & 57.8 & 58.3 & 60.3 & 58.9 & 33 & 33 & 33 & 33 \\
\bottomrule
\end{tabular}
\end{table*}

\subsection{Supplementary Analysis of Adversarial Intensity}
\label{app:intensity}

This chapter supplements the content of Section~\ref{sec:multi-turn}.
Table~\ref{tab:intensity_analysis} details the impact of adversarial intensity on model performance.
\begin{itemize}
    \item \textbf{Logic Degradation:} As expected, Logic Scores generally decrease as intensity rises from Zero to Strong (e.g., LogisticsDelivery: $84.0$ to $54.5$), confirming that adversarial user behaviors successfully challenge the agent's reasoning capabilities.
    \item \textbf{Format Error Escalation:} The JSON Error rate consistently increases with intensity (Average: $1.2\%$ to $3.7\%$), indicating that high-pressure scenarios induce ``instruction drift,'' causing models to violate output formatting constraints.
    \item \textbf{Scenario Specifics:} In some scenarios like AirlineRefund, performance actually improves from Zero to Weak, likely because the ``Zero'' setting involves trivial queries that some powerful models might over-complicate, whereas ``Weak'' scenarios provide clearer task structures.
\end{itemize}

\begin{table*}[htbp]
\centering
\caption{Adversarial Intensity Analysis: Average Performance Across All Models (0-100 Scale).}
\label{tab:intensity_analysis}

\begin{tabular}{l|ccc|ccc|ccc|ccc}
\toprule
\multirow{2}{*}{\textbf{Scenario}} & \multicolumn{3}{c|}{\textbf{Overall}} & \multicolumn{3}{c|}{\textbf{Logic}} & \multicolumn{3}{c|}{\textbf{Chat Quality}} & \multicolumn{3}{c}{\textbf{JSON Error (\%)}} \\
 & Zero & Weak & Strong & Zero & Weak & Strong & Zero & Weak & Strong & Zero & Weak & Strong \\
\midrule
AirlineRefund & 49.7 & 63.2 & 60.9 & 46.3 & 63.2 & 60.3 & 63.3 & 63.0 & 63.4 & 1.6 & 1.8 & 3.4 \\
E-commerceRefund & 73.1 & 68.8 & 58.9 & 74.8 & 69.7 & 58.1 & 66.2 & 65.2 & 62.0 & 1.0 & 1.5 & 3.9 \\
LogisticsDelivery & 78.2 & 59.9 & 56.4 & 84.0 & 59.5 & 54.5 & 55.2 & 61.6 & 63.6 & 0.6 & 2.0 & 3.8 \\
OnlineEducation & 75.7 & 74.4 & 62.6 & 78.1 & 76.7 & 62.0 & 66.1 & 65.1 & 64.9 & 2.3 & 2.9 & 4.5 \\
PropertyService & 65.1 & 61.4 & 67.0 & 64.4 & 60.3 & 68.8 & 67.9 & 65.9 & 60.0 & 0.9 & 1.0 & 3.1 \\
TelecomPackage & 66.2 & 69.5 & 54.8 & 67.7 & 72.6 & 55.2 & 60.1 & 57.1 & 53.3 & 1.0 & 1.6 & 3.1 \\
\midrule
Average & 68.0 & 66.2 & 60.1 & 69.2 & 67.0 & 59.8 & 63.1 & 63.0 & 61.2 & 1.2 & 1.8 & 3.7 \\
\bottomrule
\end{tabular}
\end{table*}

\subsection{Supplementary Analysis of Correlation Analysis of Sub-Metrics}
\label{app:submetrics}

This chapter supplements the content of Section~\ref{sec:sub_metric}.
Table~\ref{tab:subdimensions} breaks down the Logic and Chat scores into their constituent sub-dimensions.
\begin{itemize}
    \item \textbf{The Execution Gap:} A significant disparity exists between Classification Accuracy (Avg: 72.9) and Action Correctness (Avg: 41.7). This gap is most pronounced in EcommerceRefund ($88.0$ vs. $29.5$), proving that while models are adept at understanding intent, they struggle to execute the correct sequence of actions in complex SOPs.
    \item \textbf{Chat Quality Imbalance:} Within Chat Quality, Linguistic Quality (Avg: 69.7) and Instruction Compliance (Avg: 71.8) are high, whereas Anthropomorphism (Avg: 57.3) is notably lower. This suggests that current LLMs are polite and compliant but lack the ``human touch'' or empathy required for high-quality customer service.
\end{itemize}

\begin{table*}[htbp]
\centering
\caption{Sub-dimension Analysis: Average Performance Across All Models. Path: Path Correctness; Finals: Finals Correctness; Class.: Classification Accuracy; Ling.: Linguistic Quality; Anth.: Anthropomorphism; Cont.: Content Utility; Satis.: User Satisfaction; Instr.: Instruction Compliance (0-100 Scale).}
\label{tab:subdimensions}

\begin{tabular}{l|ccc|ccccc}
\toprule
\multirow{2}{*}{\textbf{Scenario}} & \multicolumn{3}{c|}{\textbf{Logic Ability}} & \multicolumn{5}{c}{\textbf{Chat Quality}} \\
 & Path & Action & Class. & Ling. & Anth. & Cont. & Satis. & Instr. \\
\midrule
AirlineRefund & 64.4 & 29.4 & 72.5 & 68.3 & 56.2 & 60.6 & 62.0 & 73.1 \\
E-commerceRefund & 66.0 & 29.5 & 88.0 & 70.1 & 58.8 & 62.1 & 64.0 & 71.4 \\
LogisticsDelivery & 64.3 & 32.3 & 66.9 & 68.9 & 55.5 & 58.6 & 60.0 & 71.6 \\
OnlineEducation & 67.0 & 43.0 & 80.2 & 71.6 & 62.3 & 60.0 & 61.7 & 73.0 \\
PropertyService & 67.2 & 57.1 & 64.0 & 71.3 & 58.3 & 60.8 & 62.1 & 75.0 \\
TelecomPackage & 74.6 & 58.7 & 65.9 & 67.9 & 52.8 & 52.1 & 53.9 & 66.3 \\
\midrule
Average & 67.3 & 41.7 & 72.9 & 69.7 & 57.3 & 59.0 & 60.6 & 71.8 \\
\bottomrule
\end{tabular}

\end{table*}

\section{Scenario Extending}
\label{app:scenario_extending}

\subsection{Scenario Configuration Template}
\label{sec:scenario_config_template}

Our benchmark includes six customer service scenarios. Each scenario follows a unified configuration structure:

\subsubsection{General Structure}
\label{subsec:scenario_structure}

Each scenario configuration contains the following components:

\begin{itemize}
\item \textbf{Scenario Metadata:}
  \begin{itemize}
  \item \texttt{scenario\_id}: Unique identifier for the scenario
  \item \texttt{scenario\_name}: Human-readable name
  \item \texttt{description}: Brief description of the scenario
  \end{itemize}

\item \textbf{Classification Fields:} Task-specific fields that the agent must classify based on dialogue context and system information. Each field has:
  \begin{itemize}
  \item \texttt{field\_name}: Name of the classification field
  \item \texttt{data\_type}: Data type (boolean, string, enum, etc.)
  \item \texttt{options}: List of valid values
  \item \texttt{description}: Detailed description of the field
  \end{itemize}

\item \textbf{System Variables:} Backend information available to the agent (e.g., user credit level, package status, order status)

\item \textbf{Actions:} Possible actions the agent can take. Each action has:
  \begin{itemize}
  \item \texttt{action\_name}: Name of the action
  \item \texttt{description}: What the action does
  % \item \texttt{parameters}: Optional parameters for the action
  \end{itemize}

\item \textbf{SOP (Standard Operating Procedure):} A decision tree that defines:
  \begin{itemize}
  \item Stage sequence (e.g., stage1, stage2, ...)
  \item Branching conditions based on classification fields and system variables
  \item Terminal actions at leaf nodes
  \end{itemize}

\item \textbf{Evaluation Metrics:}
  \begin{itemize}
  \item \textit{Path Accuracy}: Whether the agent follows the correct SOP path
  \item \textit{Action Accuracy}: Whether the agent selects the correct final action
  \item \textit{Classification Accuracy}: Accuracy of classifying dialogue fields
  \item \textit{Chat Quality}: LLM-judged quality of agent responses (5 dimensions)
  \item \textit{JSON Error Rate}: Whether the agent outputs valid JSON
  \end{itemize}
\end{itemize}

\subsubsection{Six Scenarios Overview}
\label{app:six_scenarios}

\begin{enumerate}
\item \textbf{Online Education}: Student inquiry handling, complaint resolution, and refund negotiation
\item \textbf{E-commerce Refund}: Refund processing, order issue handling, and logistics anomalies
\item \textbf{Telecom Package}: Package subscription, modification, billing inquiry, and complaints
\item \textbf{Property Service}: Fee consultation, complaint handling, and maintenance services
\item \textbf{Logistics Delivery}: Package tracking, delivery issues, loss compensation, and returns
\item \textbf{Airline Refund}: Flight rebooking/refund, complaints, and flight information inquiry
\end{enumerate}

All six scenarios follow the same configuration structure but differ in their specific fields, actions, and SOP logic.

\subsection{User Simulator Configuration Template}
\label{sec:user_config_template}

To simulate realistic customer interactions, we configure user simulators with:

\subsubsection{User Profile Components}
\label{subsec:user_profile}

\begin{itemize}
\item \textbf{User Intent}: The user's goal in the conversation (e.g., "inquiry", "complaint", "refund request")
\item \textbf{Adversarial level of intent}: Three levels to control conversation difficulty:
  \begin{itemize}
  \item \textit{Zero Adversarial Intensity}: Cooperative users who accept recommendations
  \item \textit{Weak Adversarial Intensity}: Users with mild concerns or questions
  \item \textit{Strong Adversarial Intensity}: Demanding or dissatisfied users requiring negotiation
  \end{itemize}
\item \textbf{Personality Traits}: Defines user's communication style (e.g., "friendly", "impatient", "detail-oriented")
% \item \textbf{Problem Background}: Context explaining the user's situation
% \item \textbf{Goal}: What the user wants to achieve in the conversation
\item \textbf{Initial System State}: Backend information about the user (e.g., order history, account status)
\end{itemize}

\subsubsection{Interaction Guidelines}
\label{subsec:interaction_guidelines}

User simulators follow these behavioral guidelines:

\textbf{Positive Behaviors (What users should do):}
\begin{itemize}
\item Use natural language consistent with their personality
\item Gradually disclose information over multiple turns
\item Stay focused on their intent
\item Respond appropriately to agent's actions
\item Maintain consistent emotion and adversarial level
\end{itemize}

\textbf{Negative Behaviors (What users should NOT do):}
\begin{itemize}
\item Do not switch intents mid-conversation
\item Do not reveal all information at once
\item Do not end the conversation prematurely
\item Do not break character or mention being simulated
\end{itemize}

\subsection{Example: Telecom Package Scenario}
\label{sec:telecom_example}

This subsection provides a complete example using the Telecom Package scenario to illustrate the configuration, prompts, and decision paths.

\subsubsection{Scenario Configuration}
\label{subsec:telecom_config}

\textbf{Scenario:} Telecom Package Subscription and Management

\textbf{Description:} Handling customer inquiries about package subscription, package modification, billing inquiries, and complaint processing.

\textbf{Classification Fields:}
\begin{itemize}
\item \texttt{ConsumptionType}: User's dialogue intention (Options: Enquiry, Change, Cancel)
\item \texttt{ApplicationTendency}: Whether user tends to subscribe to recommended package (Options: Agree, Reject, Hesitate)
\item \texttt{ConsumptionProfile}: Type of package user prefers (Options: Data, Voice)
\item \texttt{EmotionTag}: User's emotion in dialogue (Options: Calm, Discontent)
\end{itemize}

\textbf{System Variables:}
\begin{itemize}
\item \texttt{PackageStatus}: User's current package status (Contracted/NoContract)
\item \texttt{Penalty}: Penalty fee if user cancels contracted package (integer)
\end{itemize}

\textbf{Actions:}
\begin{itemize}
\item \texttt{ChangeOrder}: Process package change for user
\item \texttt{GoodBye}: Politely end the conversation
\item \texttt{TransHuman}: Transfer to human agent
\end{itemize}

\textbf{SOP Stages:}
\begin{enumerate}
\item \texttt{stage1}: Field Classification - Classify 4 fields based on dialogue
\item \texttt{stage2}: User Consumption Intention Judgment - Branch by ConsumptionType
\item \texttt{stage3}: User Consumption Profile Judgment - Branch by ConsumptionProfile
\item \texttt{stage4}: User Package Status Judgment - Branch by PackageStatus
\item \texttt{stage5}: Contract Penalty Situation - Branch by Penalty amount
\item \texttt{stage6}: User Application Tendency Judgment - Branch by ApplicationTendency
\item \texttt{stage7}: User Emotion Judgment - Branch by EmotionTag
\end{enumerate}

\subsubsection{Prompts}
\label{subsec:telecom_prompts}
% 通用版本（可自定义颜色和标题）

\begin{enumerate}[wide=0pt]
\item \texttt{Customer Service Agent Prompt}
\label{subsubsec:agent_prompt}
\begin{promptbox}[gray!60!black]{Agent Prompt}
% \small\ttfamily
\raggedright

You are a professional intelligent customer service representative handling [telecommunications package subscriptions]. You must process user enquiries regarding package subscriptions according to the following Standard Operating Procedure (SOP) and system variables, outputting a complete response in JSON format.

[System Variables Introduction]
PackageStatus: User package status (Contracted/NoContract)\\
Penalty: Penalty fee user needs to pay (int)

[SOP Flow Introduction]\\
1. Field Classification (stage1): Classify the following 4 fields based on the given dialogue history, then jump to stage2.\\
\hspace*{1em}- ConsumptionType: User dialogue intent (Enquiry/Change/Cancel)\\
\hspace*{1em}- ApplicationTendency: Whether user tends to apply for recommended package (Agree/Reject/Hesitate)\\
\hspace*{1em}- ConsumptionProfile: Package type user prefers (Data/Voice)\\
\hspace*{1em}- EmotionTag: User emotion in dialogue (Calm/Discontent)

2. User Consumption Intent Judgment (stage2): Jump based on [ConsumptionType] field.\\
\hspace*{1em}- Jump Logic: Based on the value of [ConsumptionType], Enquiry$\rightarrow$stage3; Change$\rightarrow$stage4; Cancel$\rightarrow$stage5.

3. User Consumption Profile Judgment (stage3): Jump based on [ConsumptionProfile] field.\\
\hspace*{1em}- Jump Logic: Based on the value of [ConsumptionProfile], jump to stage6.

4. User Package Status Judgment (stage4): Jump based on system variable [PackageStatus].\\
\hspace*{1em}- Jump Logic: Based on the value of system variable [PackageStatus], Contracted$\rightarrow$stage5; NoContract$\rightarrow$ACTION=ChangeOrder$\rightarrow$END.

5. Contract Penalty Situation (stage5): Jump based on system variable [Penalty].\\
\hspace*{1em}- Jump Logic: Based on the value of system variable [Penalty], Penalty=0$\rightarrow$ACTION=ChangeOrder$\rightarrow$END; Penalty$\neq$0$\rightarrow$stage7.

6. User Application Tendency Judgment (stage6): Jump based on [ApplicationTendency] field.\\
\hspace*{1em}- Jump Logic: Based on [ApplicationTendency] field, Agree$\rightarrow$stage4; Reject/Hesitate$\rightarrow$ACTION=GoodBye$\rightarrow$END.

7. User Emotion Judgment (stage7): Jump based on [EmotionTag] field.\\
\hspace*{1em}- Jump Logic: Based on [EmotionTag] field, Calm$\rightarrow$ACTION=ChangeOrder$\rightarrow$END; Discontent$\rightarrow$ACTION=TransHuman$\rightarrow$END.

[Action Descriptions]\\
- ChangeOrder: Change package\\
- GoodBye: Politely end the conversation\\
- TransHuman: Transfer to human agent

[Output Format Requirements]\\
You must output in the following JSON format (do not include any other text):

\{\\
\hspace*{1em}"classification\_output": \{\\
\hspace*{2em}"ConsumptionType": "Enquiry"/"Change"/"Cancel",\\
\hspace*{2em}"ApplicationTendency": "Agree"/"Reject"/"Hesitate",\\
\hspace*{2em}"ConsumptionProfile": "Data"/"Voice",\\
\hspace*{2em}"EmotionTag": "Calm"/"Discontent"\\
\hspace*{1em}\},\\
\hspace*{1em}"cot": "Briefly explain your classification reasoning and SOP flow jump logic",\\
\hspace*{1em}"now\_path": ["stage1", "stage2", "stage3", ...],\\
\hspace*{1em}"finals": \{\\
\hspace*{2em}"Action": "ChangeOrder/GoodBye/TransHuman"\\
\hspace*{1em}\},\\
\hspace*{1em}"chat": "Your friendly, professional response to the user based on Action"\\
\}

[Key Requirements - Must Follow]\\
1. Output pure JSON only, do not include any other content (such as explanations, notes, etc.).\\
2. JSON must contain the following fields:\\
\hspace*{1em}- classification\_output (object)\\
\hspace*{1em}- cot (string)\\
\hspace*{1em}- now\_path (array)\\
\hspace*{1em}- finals (object)\\
\hspace*{1em}- chat (string)\\
3. now\_path must start from "stage1" and list the stages passed in order (e.g., ["stage1", "stage2", ...]).\\
4. The chat field must be limited to 40 words. It is a complete, concise user reply. The language must be consistent with the user's language. The content must be enclosed in double quotes, and must not contain unescaped double quotes ("), backslashes (\textbackslash), square brackets ([]), etc.; if you need to quote code or special content, please describe it in text instead of directly including code.\\
5. The complete JSON should be: \{ ... \} (The outermost layer must have one and only one pair of curly braces).

\end{promptbox}

\item \texttt{User Simulator Prompt Template}
\label{subsubsec:user_prompt}

% \begin{tcolorbox}[
%     colframe=blue!75!black,
%     colback=white,
%     coltitle=white,
%     colbacktitle=blue!75!black,
%     rounded corners,
%     arc=4mm,
%     boxrule=0.8mm,
%     fonttitle=\bfseries\small,
%     title={User Simulator Prompt},
%     breakable
% ]
% \small\ttfamily
\begin{promptbox}[blue!60!black]{User Simulator Prompt}

\raggedright
You are role-playing as a telecom operator customer, preparing to consult 
with customer service or handle package-related services.

[Your Identity]\\
- User ID: \{user\_id\}\\
- Current Package: \{current\_package\}\\
- User Intent: \{user\_intent\}\\
- Adversarial level: \{adversarial\_intensity\_description\}\\
- Your Personality: \{personality\}

% [Your Problem Background]\\
% \{problem\_background\}

% [Your Goal]\\
% \{goal\}

[Interaction Requirements]\\
* What you should do:\\
- Use natural language for communication. Preferably choose English, 
  occasionally you can choose Chinese or other languages, but once you choose 
  a language, you must remain consistent throughout the conversation\\
- Respond accordingly based on the customer service representative's reply 
  and gradually advance the conversation\\
- Strictly focus on your current intent (\{user\_intent\}), do not deviate to 
  other irrelevant topics\\
- Avoid revealing all information at once, gradually disclose details to keep 
  the conversation going for multiple rounds\\
- Keep each reply concise and natural (5-15 words), simulating the rhythm of 
  a real user's conversation\\
- When customer service asks for information, provide it gradually according 
  to your personality and background, do not rush to end the conversation\\
- When the problem is not fully resolved, continue to ask for details, confirm 
  processes, or express concerns\\
- If satisfied, express thanks and confirm follow-up steps; if not satisfied, 
  continue to express your demands

* What you should NOT do:\\
- Do not cross intent boundaries: If your intent is "package inquiry", do 
  not suddenly switch to "complaint" or "query other services"\\
- Do not end the conversation too early: Do not easily say "OK thank you" 
  and end before the problem is solved, ask for details in multiple rounds\\
- Do not switch languages during the conversation: If you start with 
  English, use English throughout; if you use Chinese, use Chinese throughout\\
- Do not provide all information at once: Simulate real users' gradual 
  information disclosure\\
- Do not deviate from role settings: Strictly act according to your 
  personality and adversarial intensity\\
- Do not mention that you are AI or simulating: Be fully immersed in the 
  user role

[Important Notes]\\
- Maintain role consistency, strictly follow your intent boundaries\\
- Adopt corresponding attitudes according to your adversarial intensity\\
- Insist on your position when necessary\\
- Let the conversation continue naturally, increase interaction rounds by 
  asking, confirming, expressing emotions, etc.
\end{promptbox}

\item \texttt{Judge Prompt Template}
\label{subsubsec:judge_prompt_template}
% \label{subsec:judge_prompt}

We use LLM-as-a-judge to evaluate chat quality. The judge evaluates five dimensions:

\textbf{Chat Quality Dimensions:}
\begin{enumerate}
\item \textbf{Linguistic Quality (20\%):} Grammar, fluency, and professional language
\item \textbf{Anthropomorphism \& Emotion (25\%):} Natural human-like responses with appropriate emotional tone
\item \textbf{Content Utility (25\%):} Accuracy and relevance of information provided
\item \textbf{User Satisfaction (15\%):} Whether the response addresses user's needs
\item \textbf{Instruction Compliance (15\%):} Following SOP requirements and action descriptions
\end{enumerate}

Each dimension is scored on a 3-level scale (3/6/9 points), and the weighted sum yields a score from 0-100.

% \begin{tcolorbox}[
%     colframe=green!60!black,
%     colback=white,
%     coltitle=white,
%     colbacktitle=green!60!black,
%     rounded corners,
%     arc=4mm,
%     boxrule=0.8mm,
%     fonttitle=\bfseries\small,
%     title={Judge Prompt Template},
%     breakable
% ]
% \small\ttfamily
\begin{promptbox}[green!60!black]{Judge Prompt Template}

\raggedright
You are an expert evaluator assessing the quality of customer service 
agent responses in a telecom package scenario.

[Your Task]\\
Evaluate the agent's response based on dialogue history, user message, and 
agent's reply. Provide:\\
1. Classification of dialogue fields (ConsumptionType, ApplicationTendency, 
   ConsumptionProfile, EmotionTag)\\
2. Chat quality evaluation across five dimensions

[Chat Quality Dimensions]\\
Each dimension is scored on a 3-level scale:

1. Linguistic Quality (Weight: 20\%)\\
   - 9 points: Fluent, professional, grammatically perfect\\
   - 6 points: Generally clear with minor issues\\
   - 3 points: Poor grammar, unclear expression

2. Anthropomorphism \& Emotion (Weight: 25\%)\\
   - 9 points: Natural, empathetic, human-like interaction\\
   - 6 points: Somewhat robotic but appropriate tone\\
   - 3 points: Completely mechanical, inappropriate emotion

3. Content Utility (Weight: 25\%)\\
   - 9 points: Accurate, comprehensive, directly addresses user needs\\
   - 6 points: Partially relevant, some information missing\\
   - 3 points: Irrelevant or incorrect information

4. User Satisfaction (Weight: 15\%)\\
   - 9 points: Fully resolves user concerns, polite and helpful\\
   - 6 points: Partially addresses concerns\\
   - 3 points: Fails to address user needs, potentially frustrating

5. Instruction Compliance (Weight: 15\%)\\
   - 9 points: Perfectly follows SOP and action requirements\\
   - 6 points: Minor deviations from instructions\\
   - 3 points: Significant violations of instructions

[Output Format]\\
\{\\
\hspace*{1em}"classification": \{\\
\hspace*{2em}"ConsumptionType": "Enquiry/Change/Cancel",\\
\hspace*{2em}"ApplicationTendency": "Agree/Reject/Hesitate",\\
\hspace*{2em}"ConsumptionProfile": "Data/Voice",\\
\hspace*{2em}"EmotionTag": "Calm/Discontent"\\
\hspace*{1em}\},\\
\hspace*{1em}"chat\_quality\_dimensions": \{\\
\hspace*{2em}"linguistic\_quality": 3 or 6 or 9,\\
\hspace*{2em}"anthropomorphism\_emotion": 3 or 6 or 9,\\
\hspace*{2em}"content\_utility": 3 or 6 or 9,\\
\hspace*{2em}"user\_satisfaction": 3 or 6 or 9,\\
\hspace*{2em}"instruction\_compliance": 3 or 6 or 9\\
\hspace*{1em}\},\\
\hspace*{1em}"classification\_reasoning": "Explanation for classification decisions",\\
\hspace*{1em}"chat\_quality\_reasoning": "Explanation for each dimension's score\\
\hspace*{7em}(e.g., 9 points means..., 6 points means...,\\
\hspace*{7em}3 points means...)"\\
\}

[Requirements]\\
1. Only output valid JSON\\
2. Scores must be exactly 3, 6, or 9 for each dimension\\
3. Provide clear reasoning for both classification and quality evaluation
\end{promptbox}

\subsubsection{Decision Paths (PathList)}
\label{subsec:telecom_pathlist}

The PathList enumerates all valid SOP decision paths for the scenario. Each path specifies:
\begin{itemize}
\item \textbf{Classification Items}: Values of classification fields
\item \textbf{System Variables}: Backend state (e.g., PackageStatus, Penalty)
\item \textbf{Expected Path}: Sequence of SOP stages
\item \textbf{Final Output}: Terminal action
\end{itemize}

\subsubsection{Example Paths}
\label{subsubsec:example_paths}

Below are three representative paths from the Telecom Package scenario:

% \begin{tcolorbox}[
%     colframe=orange!75!black,
%     colback=white,
%     coltitle=white,
%     colbacktitle=orange!75!black,
%     rounded corners,
%     arc=4mm,
%     boxrule=0.8mm,
%     fonttitle=\bfseries\small,
%     title={Path 1: Enquiry + Agree + Data + NoContract $\rightarrow$ ChangeOrder},
%     breakable
% ]
% \small\ttfamily
\begin{promptbox}[orange!75!black]{Path 1: Enquiry + Agree + Data + NoContract $\rightarrow$ ChangeOrder}
\{\\
\hspace*{1em}"Classification\_items": ["Enquiry", "Agree", "Data", "Calm"],\\
\hspace*{1em}"system\_variables": \{"PackageStatus": "NoContract", "Penalty": 0\},\\
\hspace*{1em}"expected\_path": ["stage1", "stage2", "stage3", "stage6", "stage4"],\\
\hspace*{1em}"final\_output": \{"Action": "ChangeOrder"\}\\
\}
\end{promptbox}

\begin{promptbox}[orange!75!black]{Path 2: Change + Contracted + Penalty>0 + Discontent $\rightarrow$ TransHuman}
% ,
%     breakable
% ]
% \small\ttfamily
\{\\
\hspace*{1em}"Classification\_items": ["Change", "Agree", "Data", "Discontent"],\\
\hspace*{1em}"system\_variables": \{"PackageStatus": "Contracted", "Penalty": 100\},\\
\hspace*{1em}"expected\_path": ["stage1", "stage2", "stage4", "stage5", "stage7"],\\
\hspace*{1em}"final\_output": \{"Action": "TransHuman"\}\\
\}
\end{promptbox}

% \begin{tcolorbox}[
%     colframe=orange!75!black,
%     colback=white,
%     coltitle=white,
%     colbacktitle=orange!75!black,
%     rounded corners,
%     arc=4mm,
%     boxrule=0.8mm,
%     fonttitle=\bfseries\small,
%     title=
    \begin{promptbox}[orange!75!black]{Path 3: Enquiry + Reject + Voice $\rightarrow$ GoodBye}
%     ,
    
%     breakable
% ]
% \small\ttfamily
\{\\
\hspace*{1em}"Classification\_items": ["Enquiry", "Reject", "Voice", "Calm"],\\
\hspace*{1em}"system\_variables": \{"PackageStatus": "NoContract", "Penalty": 0\},\\
\hspace*{1em}"expected\_path": ["stage1", "stage2", "stage3", "stage6"],\\
\hspace*{1em}"final\_output": \{"Action": "GoodBye"\}\\
\}
\end{promptbox}

\textbf{Note:} The complete Telecom Package scenario has 36 valid paths in total. The other five scenarios have similar PathList structures with varying numbers of paths based on their complexity.
\end{enumerate}

\section{SOP Graph of 6 Scenarios in Our Evaluation}
\label{sec:six_scenarios}

In this section, we visualize the Standard Operating Procedures (SOPs) formalized as directed graphs for the six industrial scenarios evaluated in SAGE. These graphs define the explicit logical constraints, state transitions, and action spaces that serve as the ground truth for our Rule Engine evaluation.

\textbf{1. Ecommerce Refund (ER):} 
As illustrated in Figure~\ref{fig:sop:ecommerce_refund}, this is the most complex scenario in our benchmark. The graph features a deep decision tree with intricate branching based on \texttt{ShippingStatus} (Shipped/Unshipped/Signed) and \texttt{Responsibility} attribution. Crucially, it incorporates dynamic checks on user \texttt{CreditLevel} to determine immediate refund eligibility. The agent must navigate this multi-branch logic to verify eligibility and negotiate terms, validating its capability in handling high-complexity disputes.

\textbf{2. Logistics Delivery (LD):} 
Shown in Figure~\ref{fig:sop:logistics_delivery}, this scenario centers on supply chain exception handling. The SOP graph requires the agent to track package status and process insurance claims based on \texttt{ComplaintValidity}. Key logic nodes include \texttt{RiskStatus} (intercepting risky orders) and \texttt{EmergencyLevel} (prioritizing urgent registrations), testing the agent's proficiency in managing urgency and strictly following insurance protocols.

\textbf{3. Telecom Package (TP):} 
Depicted in Figure~\ref{fig:sop:telecom_package}, this is a standardized scenario with relatively linear logic. The flow focuses on billing inquiries and plan upgrades, guided by \texttt{ConsumptionType} (Enquiry/Change/Cancel). The agent must check \texttt{PackageStatus} and \texttt{Contracted} states before executing changes. The graph primarily evaluates the agent's accuracy in instruction-following and executing upselling protocols within a structured framework.

\textbf{4. Property Service (PS):} 
As seen in Figure~\ref{fig:sop:property_service}, this scenario emphasizes community coordination. The graph divides flows into Payment, Complaint, and Repair, requiring checks on \texttt{HouseStatus} (Occupied/Vacant) and \texttt{FeePaymentStatus}. The logic is designed to test the agent's ability to coordinate offline services (e.g., scheduling repairs) while managing resident emotions in noise complaint sub-branches.

\textbf{5. Airline Refund (AR):} 
Illustrated in Figure~\ref{fig:sop:airline_refund}, this is a high-complexity, high-pressure scenario governed by rigid policies. The graph enforces strict validation of \texttt{ChangeReason} (Personal vs. Airline/Weather) and \texttt{memberLevel} (VIP/Regular). The agent must precisely calculate dynamic cancellation fees based on these variables. This structure rigorously tests the agent's precision in policy adherence and its ability to de-escalate passenger anxiety under time constraints.

\textbf{6. Online Education (OE):} 
Shown in Figure~\ref{fig:sop:online_education}, this scenario focuses on rigorous \textbf{Risk Control}. The graph explicitly includes a \texttt{isRiskUser} check to identify potential malicious refunders based on historical records. The agent must strictly adhere to complex refund formulas and select specific \texttt{PLAN} options (A-F) based on the user's learning dependency. This design serves as a stress test for logical reasoning and compliance with anti-fraud protocols.

% 第一页（2×2布局）
\begin{figure*}[htbp]
  \centering
  
  % 第一行
  \begin{subfigure}{0.48\textwidth}
    \centering
    \includegraphics[width=0.9\textwidth]{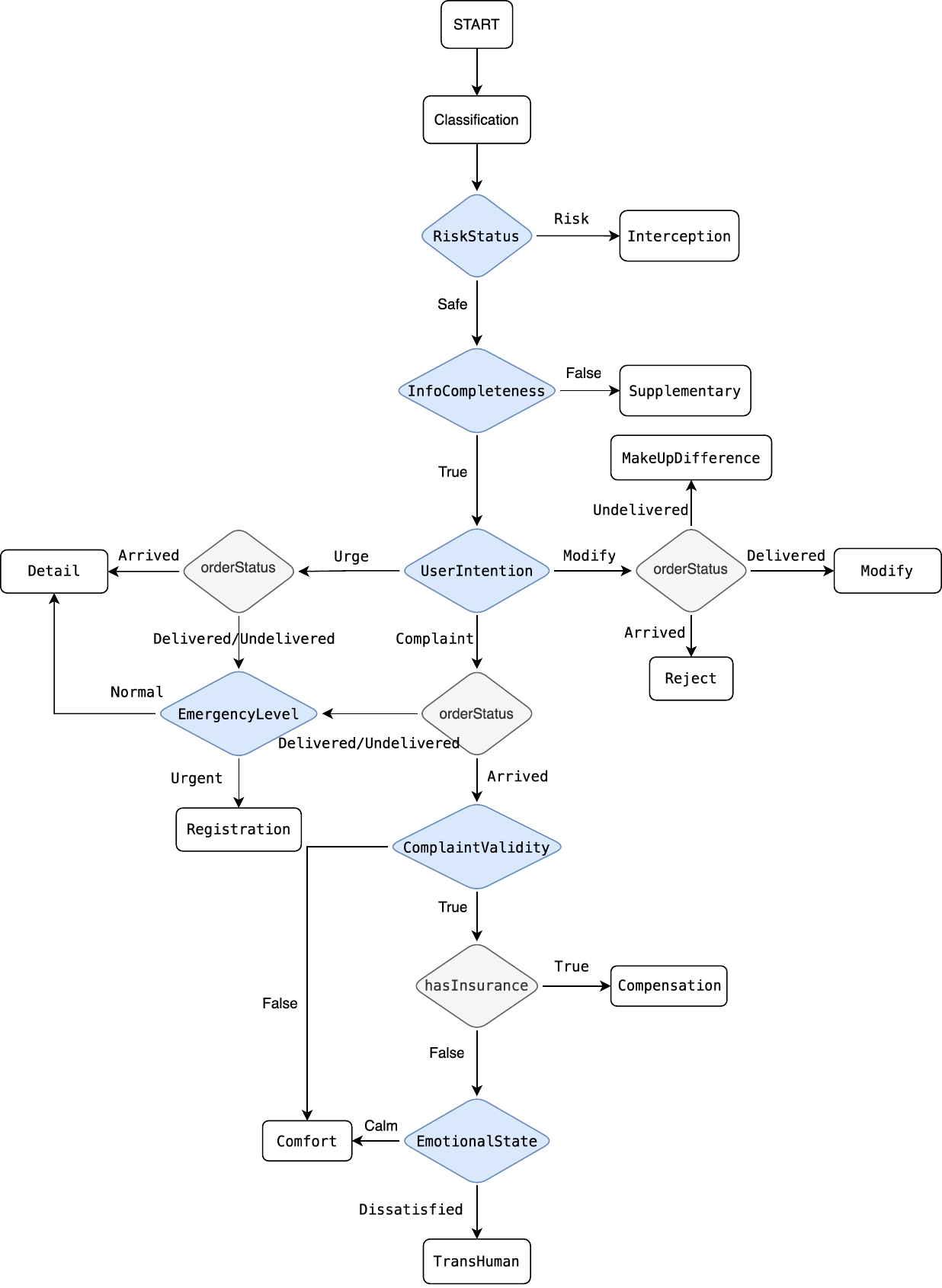}
    \caption{Logistics Delivery (LD) SOP}
    \label{fig:sop:logistics_delivery}
  \end{subfigure}
  \hfill
    \begin{subfigure}{0.48\textwidth}
    \centering
    \includegraphics[width=0.9\textwidth]{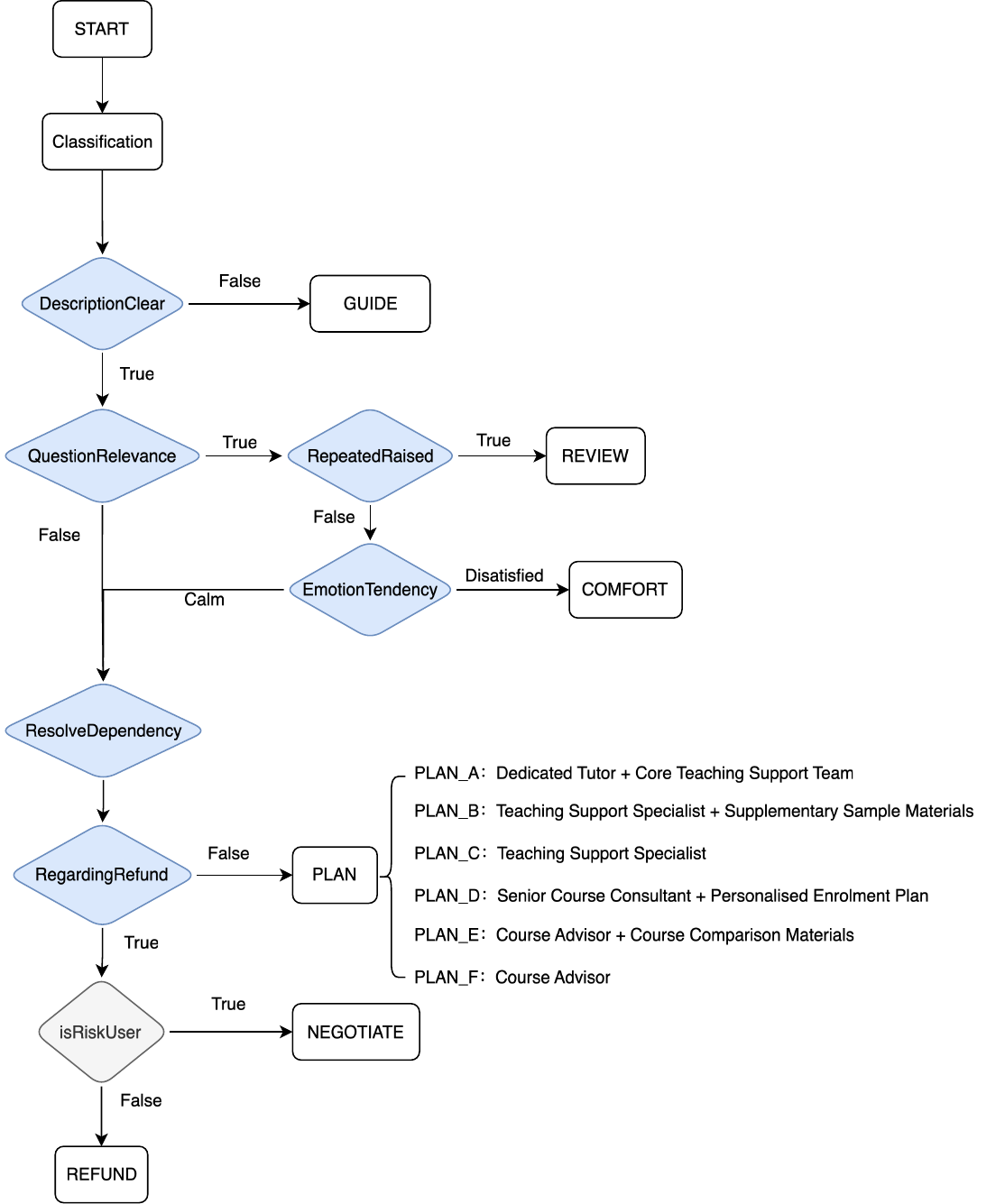}
    \caption{Online Education (OE) SOP}
    \label{fig:sop:online_education}
  \end{subfigure}

  % 第二行
  \begin{subfigure}{0.48\textwidth}
    \centering
    \includegraphics[width=0.8\textwidth]{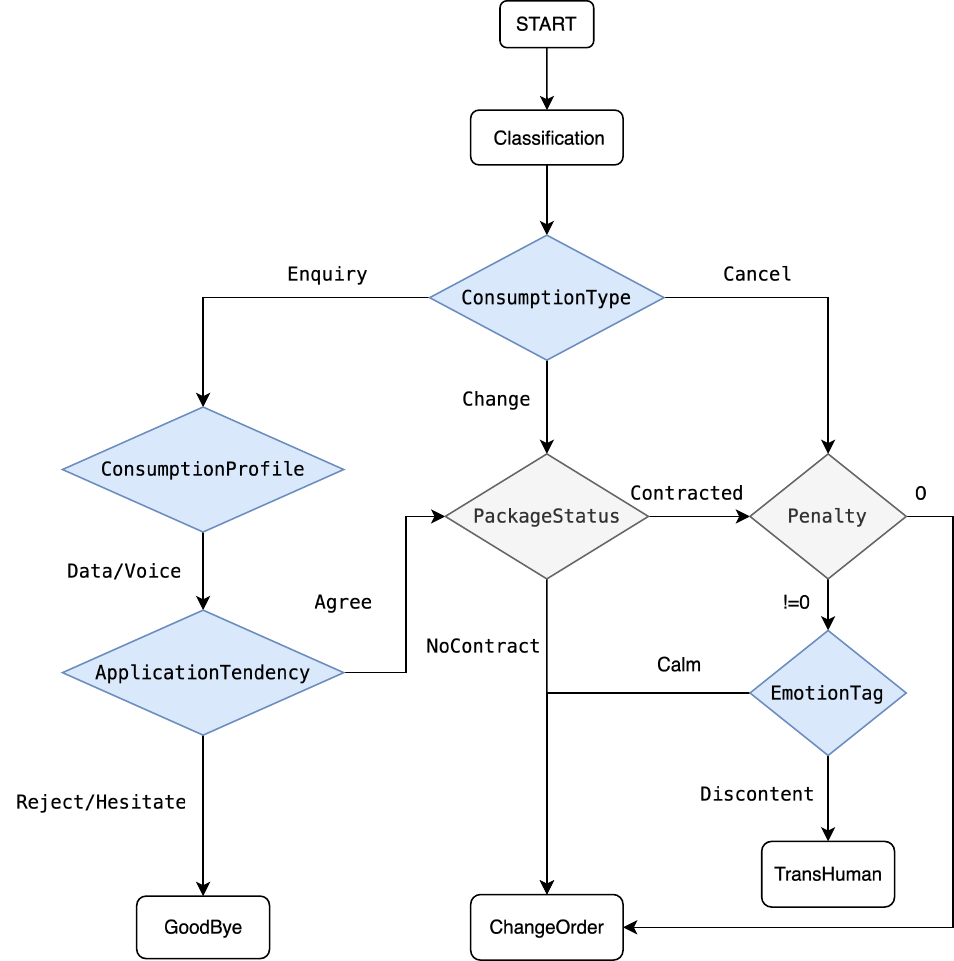}
    \caption{Telecom Package (TP) SOP}
    \label{fig:sop:telecom_package}
  \end{subfigure}
  \hfill
  \begin{subfigure}{0.48\textwidth}
    \centering
    \includegraphics[width=0.85\textwidth]{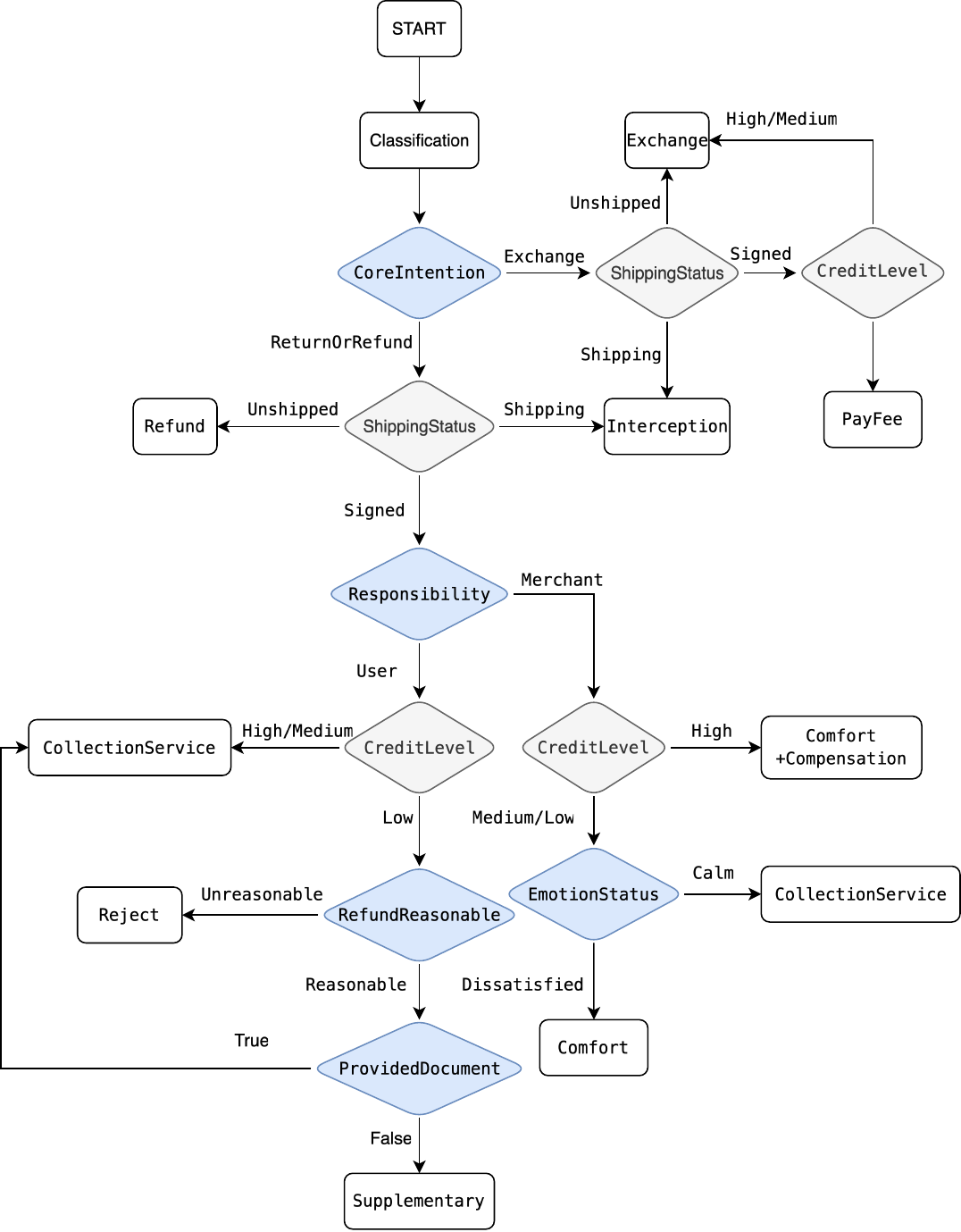}
    \caption{E-commerce Refund (ER) SOP}
    \label{fig:sop:ecommerce_refund}
  \end{subfigure}
  \caption{Standard Operating Procedures (SOPs) for Six Industrial Scenarios evaluated in SAGE. These directed graphs define the logical constraints and action spaces for each domain.(continued)}
  \label{fig:sop_group_1}
\end{figure*}

% 第二页（2行,每行1张图）
\begin{figure*}[htbp]
  \ContinuedFloat
  \centering
  
  % 第一行
  \begin{subfigure}{0.6\textwidth}
    \centering
    \includegraphics[width=\textwidth]{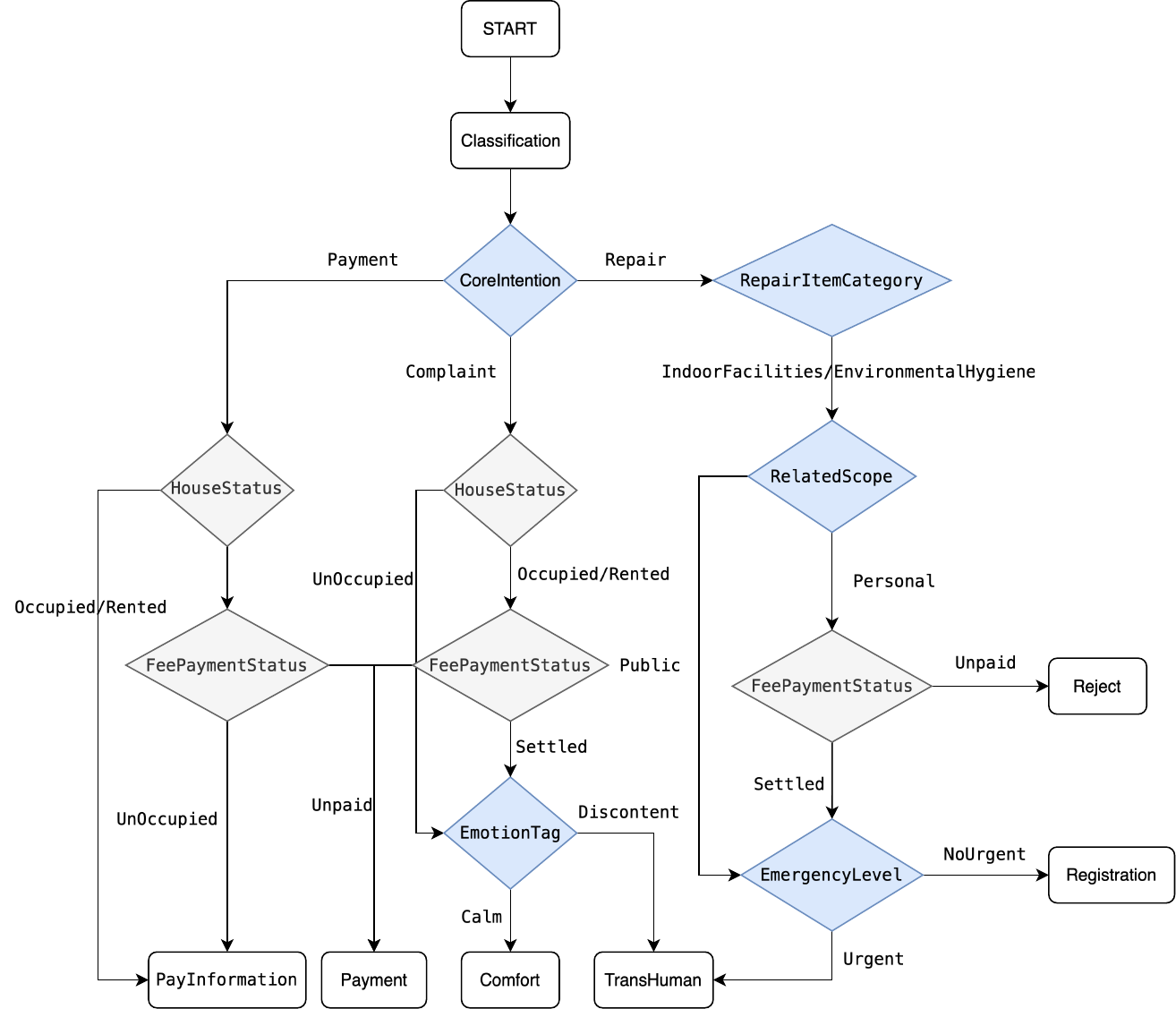}
    \caption{Property Service (PS) SOP}
    \label{fig:sop:property_service}
  \end{subfigure}

  % 第二行
  \begin{subfigure}{0.7\textwidth}
    \centering
    \includegraphics[width=\textwidth]{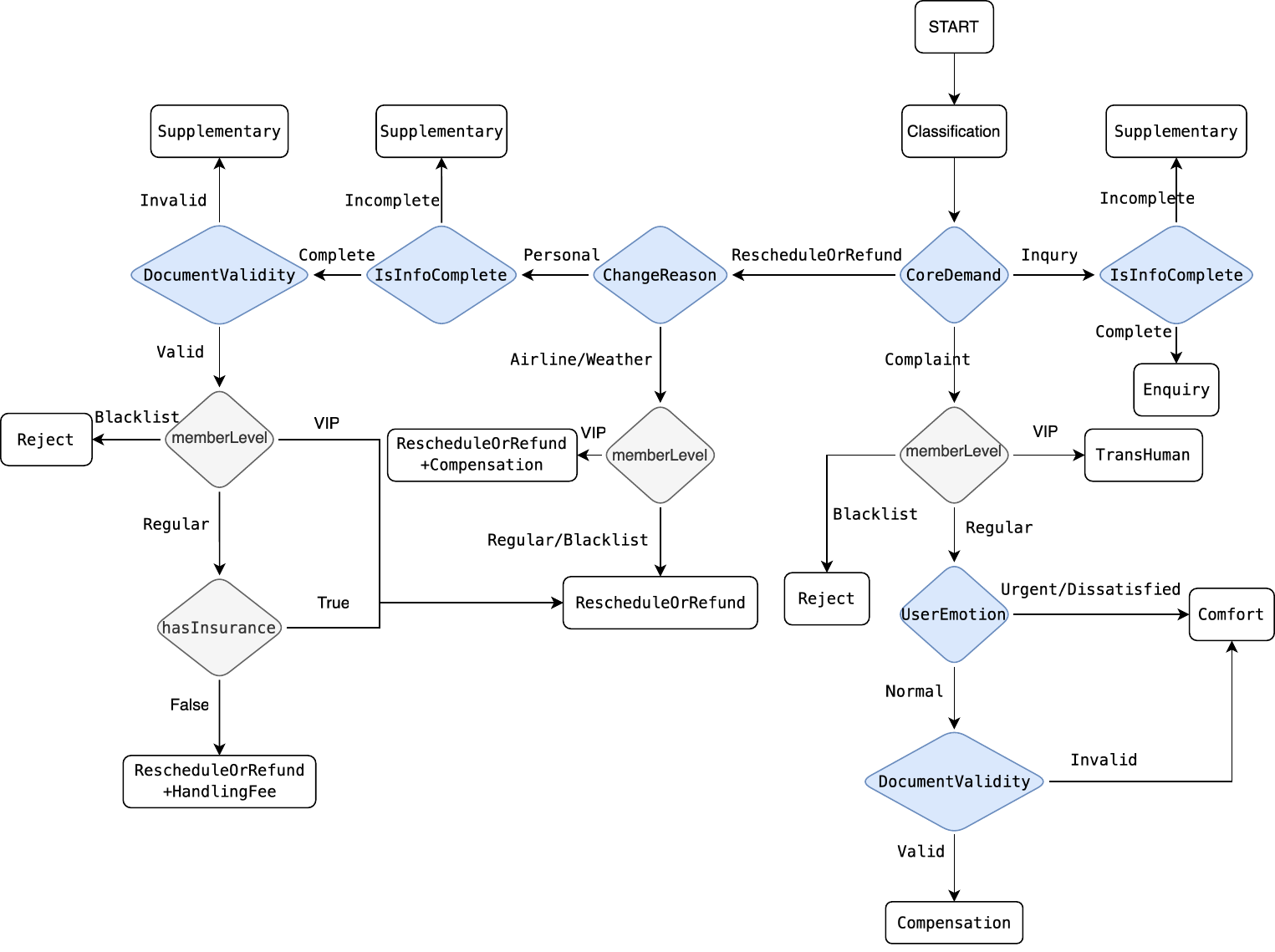}
    \caption{Airline Refund (AR) SOP}
    \label{fig:sop:airline_refund}
  \end{subfigure}
  
  \caption{Standard Operating Procedures (SOPs) for Six Industrial Scenarios.}
  \label{fig:sop_group_2}
\end{figure*}

\end{document}